\documentclass[conference]{IEEEtran}
\IEEEoverridecommandlockouts
\usepackage{cite}
\usepackage{amsmath,amssymb,amsfonts}
\usepackage{algorithmic}
\usepackage{graphicx}
\usepackage{textcomp}
\usepackage{xcolor}
\usepackage{times}

\usepackage{subcaption}
\usepackage{multirow}
\usepackage{makecell}
\usepackage{mathtools}
\usepackage{dsfont}
\usepackage{float}
\usepackage{siunitx} 
\usepackage{xcolor}
\usepackage{url}
\usepackage{bbm}
\usepackage{soul} 
\usepackage[makeroom]{cancel}
\usepackage{booktabs}

\def\BibTeX{{\rm B\kern-.05em{\sc i\kern-.025em b}\kern-.08em
    T\kern-.1667em\lower.7ex\hbox{E}\kern-.125emX}}
\begin{document}

\title{ANYmal Parkour: Learning Agile Navigation for Quadrupedal Robots} 

\author{
    \IEEEauthorblockN{David Hoeller}
    \IEEEauthorblockA{\textit{ETH Zurich} \\
    \textit{NVIDIA}\\
    dhoeller@ethz.ch\\
    Equal contribution}
    \and
    \IEEEauthorblockN{Nikita Rudin}
    \IEEEauthorblockA{\textit{ETH Zurich} \\
    \textit{NVIDIA}\\
    rudinn@ethz.ch\\
    Equal contribution}
    \and
    \IEEEauthorblockN{Dhionis Sako}
    \IEEEauthorblockA{\textit{ETH Zurich} \\
    dsako@ethz.ch}
    \and
    \IEEEauthorblockN{Marco Hutter}
    \IEEEauthorblockA{\textit{ETH Zurich} \\
    mahutter@ethz.ch}
}

\maketitle

\begin{abstract}
Performing agile navigation with four-legged robots is a challenging task due to the highly dynamic motions, contacts with various parts of the robot, and the limited field of view of the perception sensors. 
In this paper, we propose a fully-learned approach to train such robots and conquer scenarios that are reminiscent of parkour challenges. The method involves training advanced locomotion skills for several types of obstacles, such as walking, jumping, climbing, and crouching, and then using a high-level policy to select and control those skills across the terrain. Thanks to our hierarchical formulation, the navigation policy is aware of the capabilities of each skill, and it will adapt its behavior depending on the scenario at hand. Additionally, a perception module is trained to reconstruct obstacles from highly occluded and noisy sensory data and endows the pipeline with scene understanding. 
Compared to previous attempts, our method can plan a path for challenging scenarios without expert demonstration, offline computation, a priori knowledge of the environment, or taking contacts explicitly into account.
While these modules are trained from simulated data only, our real-world experiments demonstrate successful transfer on hardware, where the robot navigates and crosses consecutive challenging obstacles with speeds of up to two meters per second. The supplementary video can be found on the project website: https://sites.google.com/leggedrobotics.com/agile-navigation
\end{abstract}


\section{Introduction}
\footnote{Under Review}
Parkour, also known as free-running, is a discipline originating in the late 80s that has gained in popularity with the advent of the internet. 
Free-runners perform acrobatic stunts where the goal is to attain a hard-to-reach location in the most elegant and efficient manner. 
It involves navigating through the environment by walking, running, climbing, and jumping over obstacles, and the athlete must coordinate these agile skills in a precisely-timed sequence.
This discipline requires years of practice to develop the necessary competencies, intuitions, and reflexes and is considered particularly dangerous.

While legged robots aspire to be as nimble and agile as humans or animals, we are still far from fully exploiting their capabilities to achieve similar behaviors. By aiming to match the agility of free-runners, we can better understand the limitations of each component in the pipeline from perception to actuation, circumvent those limits, and generally increase the capabilities of our robots which in return paves the road for many new applications such as search and rescue in collapsed buildings or complex natural terrains.

In such scenarios, the robot must sense its environment to develop an understanding of the rapidly changing surrounding scene and select a feasible path and sequence of motions based on its set of skills. In the case of large and challenging obstacles, it has to perform dynamic maneuvers at the limits of actuation while accurately controlling the motion of the base and limbs.
All of the above must be achieved in real-time with limited onboard computing and using its exteroceptive sensors' partial and noisy information.

The complexity of the task exacerbates many of the challenges commonly faced by mobile robots:

\begin{itemize}
    \item The locomotion controller cannot rely on a stable and periodic gait but must use completely different motions and make contact with its limbs depending on the obstacles at hand.
    \item State estimation is prone to heavy drift due to high-impact forces and contacts on various parts of the robot.
    \item Perceiving the environment is difficult since self-occlusions and the limited field of view of the sensors result in a partial view of the scene.
    \item The planner has to reason about the environment, understand the kinematic and dynamic capabilities of the robot and know the limitations of its low-level controllers to produce feasible trajectories.
    \item Any latency in the system can be catastrophic during fast motions. As such, the processing of sensory data and the inference of controllers must be performed with minimal delays.
\end{itemize}

\begin{figure*}[pt!]
\centering
\includegraphics[width=0.95\textwidth]{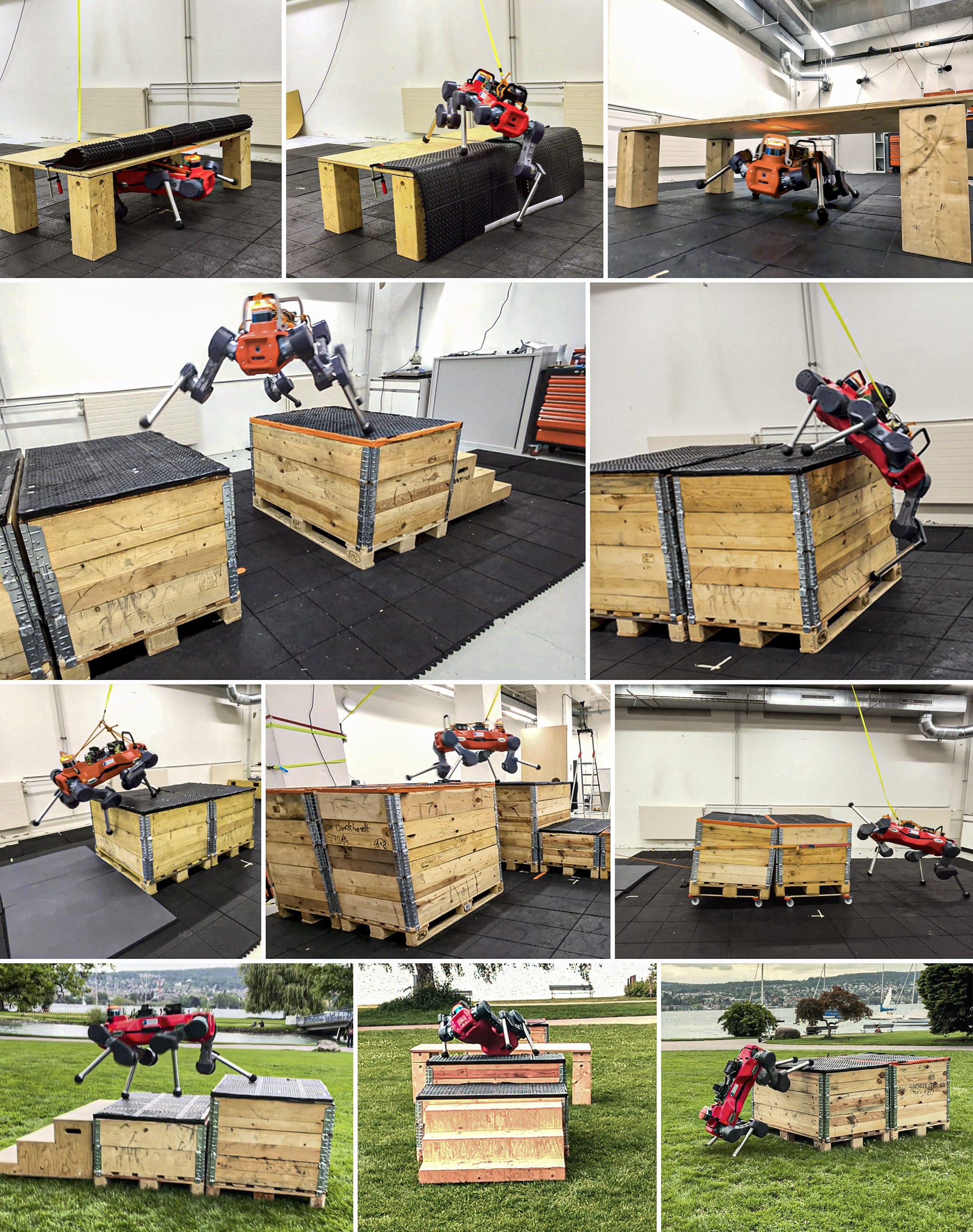}
\caption[Trajectories of the robot in simulation.]{Deployment of the pipeline on the quadrupedal robot ANYmal D. The robot performs highly dynamic maneuvers and makes contacts with its limbs where necessary.}
\label{fig:collage}
\end{figure*}

\subsection{Method overview}
\begin{figure*}[t!]
\centering
\includegraphics[width=\textwidth]{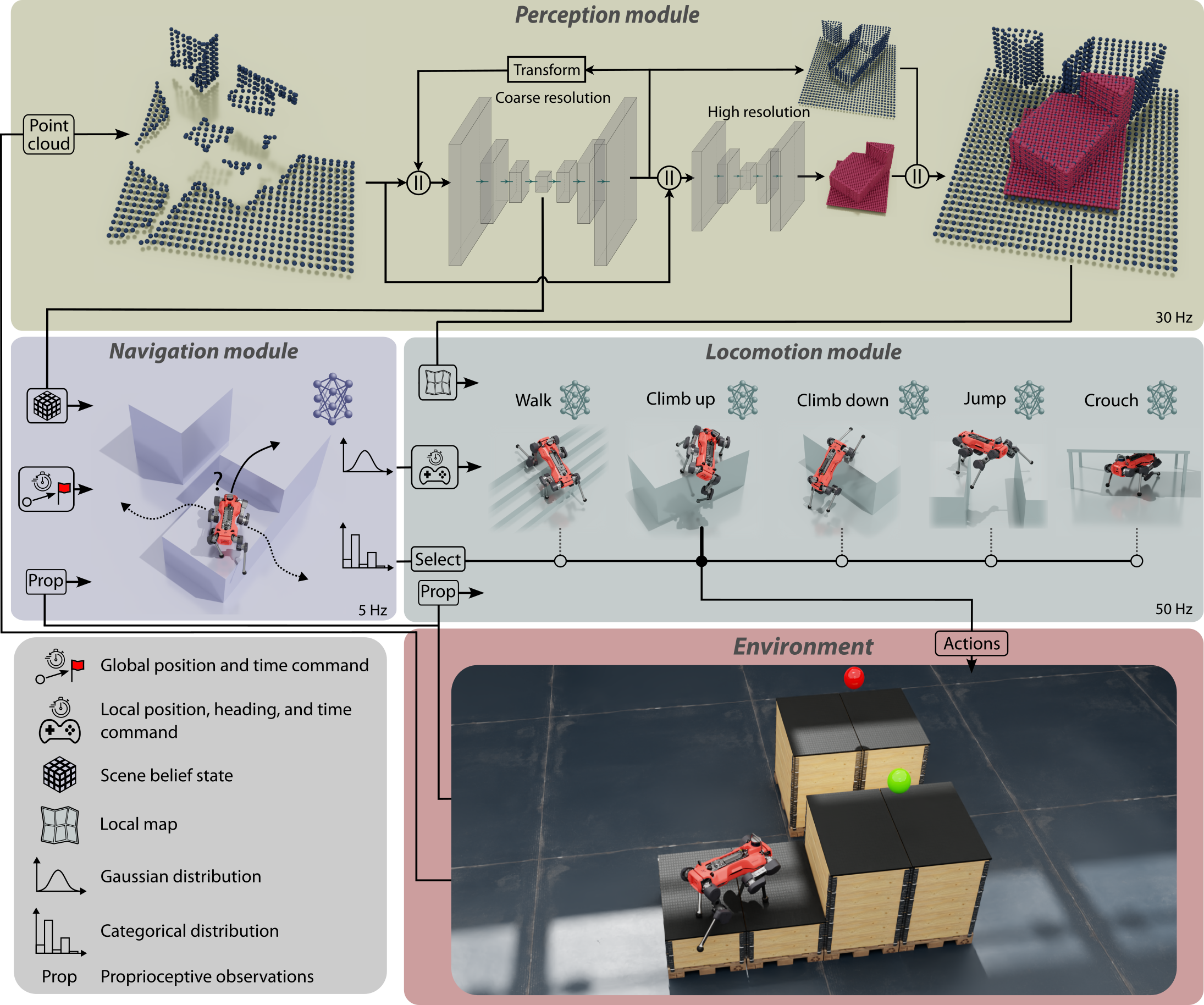}
\caption{Description of our approach. We decompose the problem into three components: The perception module receives the point cloud measurements to estimate the scene's layout and produces a latent tensor and a map. The locomotion module contains several low-level skills that can overcome specific scenarios. The navigation module is given a target goal and uses the latent to plan a path and select the correct skill.}
\label{fig:architecture}
\end{figure*}

This work aims to solve the above-mentioned challenges and proposes a method to perform agile navigation with a quadrupedal robot in parkour-like settings (Fig.~\ref{fig:collage}). Based on state-of-the-art methods for mapping, planning, and locomotion, the robot is trained to navigate and locomote in an environment to reach a specific target location. Limiting the allocated time forces the robot to overcome the course at high speeds and demonstrate fast decision-making.  \\
We split the pipeline into three interconnected components: a perception module, a locomotion module, and a navigation module (Fig.~\ref{fig:architecture}). The perception module receives point cloud measurements from the onboard cameras and the LiDAR and computes an estimate of the terrain around the robot, as well as a compact latent vector that represents the belief state of the scene. The locomotion module contains a catalog of locomotion skills that can overcome specific terrains. For this work, we train five policies that can walk on irregular terrain, jump over gaps, climb up and down high obstacles, and crouch in narrow passages. Using the latent tensor of the perception module, the navigation module guides the locomotion module in the environment by selecting which skill to activate and providing intermediate commands. Each of these learning-based modules is trained in simulation. 

We design randomized obstacle fields containing stairs, inclined surfaces, boxes, gaps, and tables. While the locomotion policies are trained on single obstacles, the perception and navigation modules are trained on various arrangements of these obstacles.

Finally, after training in simulation, all the modules are deployed in the real world.

\subsection{Contributions}
In our experimental validation, we demonstrate the system's ability to solve the problem autonomously, resulting in behaviors not shown before with such platforms. The robot can cross difficult terrains with speeds of up to \SI{2}{\meter/\second} and make the right navigation decisions to reach the target in time.
The locomotion controllers perform precise and agile movements, sometimes on narrow boxes barely the size of the robot's footprint, and leverage the system's full range of motion to pass higher obstacles. The mapping pipeline, which provides updates at a high frequency, correctly reconstructs the scene despite state estimation and sensing noise stemming from the robot's fast speeds. Finally, the planner uses the available information and its intrinsic knowledge of each skill's capabilities to guide the robot around the course on a feasible path. All of these components are designed with efficiency in mind. They scale properly when training with thousands of agents in simulation and operate in real-time on the real robot. We show that the complete pipeline can be deployed sim-to-real achieving high agility despite the harsh conditions of the real world.
We can summarize our contributions as follows:

\begin{enumerate}
    \item We propose a novel learned navigation approach that uses the belief state of the terrain reconstruction network to plan a path through intricate scenes while selecting from a library of locomotion skills. 
    Thanks to the simple architecture, inference is in the order of milliseconds.
    We modify the PPO~\cite{schulman2017ppo} algorithm to have a hybrid actor output with a Gaussian distribution for the low-level commands and a categorical distribution for skill selection.
    \item We train new and more capable locomotion skills by extending the position-based formulation in \cite{rudinAdvancedSkills}. We define new terrains, include a heading command, and use symmetry augmentation to increase the policies' performance.
    \item We develop a neural terrain reconstruction method that can handle the challenging conditions of our task. We augment the approach described in \cite{hoeller2022TREP} with a multi-resolution scheme to combine precise reconstruction near the robot with a coarse larger-scale map to have a larger view of the scene. We also modify the network architecture to allow for efficient inference with large batch sizes during RL training.
    We demonstrate the applicability of the method to complex scenes with overhanging obstacles.
    \item We deploy all modules in the real world on the ANYmal~D robot. We test the capabilities of the system on a variety of obstacle arrangements in both indoor and outdoor settings.
\end{enumerate}

\subsection{Related work}

\paragraph{Locomotion}
Perceptive legged locomotion has been tackled by multiple approaches ranging from model-based to fully learned techniques. Model predictive control (MPC) can be used to traverse challenging terrains requiring precise foot placement \cite{grandia2022perceptive, jenelten2022Tamols, kim2020loco}, but the approach is limited by the underlying model. MPC tends to fail in cases of slippage or imperfect terrain perception. Furthermore, current MPC approaches make strong assumptions about the contact schedule of the feet, proscribing any other contact between the robot and its environment.
Deep reinforcement learning has proven to be an effective solution for robust perceptive locomotion~\cite{miki2021locomotion, loquercio2022learn, gangapurwala2020RLOCTL}. Nevertheless, it is still far from exploiting its potential.

Agile locomotion has been of strong interest since the first legged robots \cite{raibert1986legged}. More recently, the research continued with quadrupedal robots running \cite{kim2019loco} and jumping over obstacles \cite{park2021jumping, nguyen2019Jumping}. In recent years, with the combined democratization of commercially available quadrupedal platforms and openly available deep reinforcement learning frameworks, various new tasks have been demonstrated. Notable examples include jumping and climbing \cite{rudinAdvancedSkills}, performing cat-like landing motions \cite{rudin2021cat, jeon2021cat}, recovering from falls \cite{hwangbo2019learning, ma2023almarecovery}, and dribbling with a football \cite{bohez2022imitate, ji2023dribblebot}.

In parallel, bipedal robots have also demonstrated their agile capabilities by walking blindly on rough terrain \cite{siekmann2021cassieblind} and jumping on obstacles \cite{li2023cassiejump}.

\paragraph{Navigation and Hierarchical Learning}
Navigation is typically achieved with a hierarchical set-up, where a planner computes a feasible and collision-free path, which a controller then tracks.
While sampling-based methods are commonly used to create such a path \cite{karaman2011RRT}, employing such techniques with legged robots is challenging due to the system's complex and hybrid nature. The robot must constantly make and break contact with the environment to influence its motion, which leads to the combinatorial explosion of the set of possible solutions. As a result, the problem is usually simplified to keep it tractable. In \cite{Tonneau2018Planning}, a solution is proposed to plan maneuvers in challenging environments for several legged robots. A path is first sampled using primitive collision shapes, and a sequence of contacts that are statically stable is planned. The procedure takes a few seconds to converge, requires a priori knowledge of the environment, and results in statically stable motions, making it infeasible for our task. The authors in \cite{wellhausen2021Nav} propose a real-time capable approach to plan a path on rough terrain with a quadrupedal robot. They constrain the solution space by estimating appropriate footholds from an elevation map but also simplify the problem by using a primitive robot morphology and assuming that the robot is always in contact with its feet. 
More related to parkour, \cite{kim2020loco} deploys a model-based system to walk on rough terrain and jump on boxes. However, the plan is computed offline, the switch to the jumping controller is hard-coded, and the system can only overcome obstacles of \SI{0.1}{\meter} with speeds of \SI{0.33}{\meter/\second}.

Arguably, learning-based methods can break down such complexity and provide a more straightforward way to guide the robot from a point A to B. Previous works have trained navigation policies from expert demonstration \cite{pfeiffer2017perception, Kaufmann2018DeepDR}, using reinforcement learning \cite{SadeghiL17, Sadeghi19, hoeller2021DeepNav}, or fully self-supervised \cite{Badgr}. 
For legged robots, the authors of \cite{bowen2021Navigation} proposed to combine sampling-based planning with a learned motion cost for global path planning, resulting in a planner aware of the underlying controller's capabilities. Unfortunately, the method requires access to a global map beforehand and operates on elevation maps, meaning that the resulting plan cannot pass underneath obstacles.
Recently, \cite{caluwaerts2023barkour} demonstrated that a quadrupedal robot can solve an obstacle course inspired by dog agility competitions using a hierarchical learning approach. Despite the promising results and the close similarity to our method, this work requires human-designed path and skill selection and is limited to a single pre-mapped environment with a motion capture system.
To the best of our knowledge, we propose the first system that can perform agile navigation with a quadrupedal robot in such challenging scenarios without a priori planning or mapping.

Hierarchical reinforcement learning has gained attention in the field of robotics as it enables robots to acquire, combine, and reuse versatile skills in order to solve complex tasks. Pre-training low-level skills with imitation learning and then controlling them through latent actions has been proposed for both character animation \cite{peng2022ASE} and robotics \cite{bohez2022imitate}. Combining multiple expert policies has also been explored by switching between policies trained to imitate fragments of motions \cite{merel2018hierarchical} or by fusing locomotion policies with gating neural networks \cite{yang2020MultiExpertLoco}.

In this work, we train locomotion skills using the position-based task formulation of \cite{rudinAdvancedSkills}. Similar to \cite{merel2018hierarchical}, the navigation module then learns to steer and switch between those skills. 

\paragraph{Perception for navigation and locomotion} 
Navigation and locomotion pipelines for legged robots heavily rely on elevation maps ~\cite{chavez2018traversability, bowen2021Navigation, miki2021locomotion, grandia2022perceptive, gangapurwala2020RLOCTL}. 
However, noise and inaccurate state estimation lead to unclean maps. To overcome these limitations, the authors in~\cite{miki2021locomotion} use a teacher-student set-up to train locomotion policies, where the student learns to deal with mapping inaccuracies. 
The approach in~\cite{miki2022EM} improves standard elevation mapping~\cite{Fankhauser2018ProbabilisticTerrainMapping} by adding various filtering operations and post-processing steps, which can explicitly realign the map to compensate for drift in the z-direction, and performs additional visibility checks to clean up outliers. We compare the reconstructions against this method in our experiments.
Elevation maps, however, have drawbacks that limit their deployment for our task: They cannot represent the full 3D configuration of the world and cannot extrapolate beyond visible data, which is necessary to pass below obstacles or to reconstruct the top surfaces of higher obstacles.
For navigation, signed distance fields ~\cite{oleynikova2017voxblox} are commonly used since they can easily be integrated into the problem formulation to avoid elements in the scene.
While these approaches produce a separate representation, the exteroceptive measurements can also be directly provided as input to the policy \cite{agarwal2022legged, loquercio2022learn}. These methods, however, involve multiple stages that provide direct supervision to the perceptive part.
In this work, we take inspiration from~\cite{hoeller2022TREP} to reconstruct the environment in 3D from point cloud data. We augment the method with a multi-resolution scheme to have a higher resolution near the robot and a lower resolution further away to have a larger view of the scene.
 
\section{Results}
We deploy the pipeline on the quadrupedal robot ANYmal~D. It weighs around \SI{55}{\kilo\gram} and has 12 series elastic actuators capable of producing a torque of \SI{85}{\newton\meter} each. To perceive the environment, it is equipped with a total of six Intel Realsense depth cameras (two in the front, two in the back, one left, one right), and a Velodyne Puck LiDAR. The whole system is implemented in several ROS nodes across different onboard computers. The locomotion and navigation modules operate synchronously in a single node on the onboard computer. The perception module is implemented on an NVIDIA Jetson Orin and operates asynchronously with the rest of the system, i.e., the navigation and locomotion policies take the last received message from the perception module to infer their respective networks. The supplementary video summarizes the proposed approach and shows indoor and outdoor experiments on the real robot.

\begin{figure*}[pt!]
\centering
\includegraphics[width=0.9\textwidth]{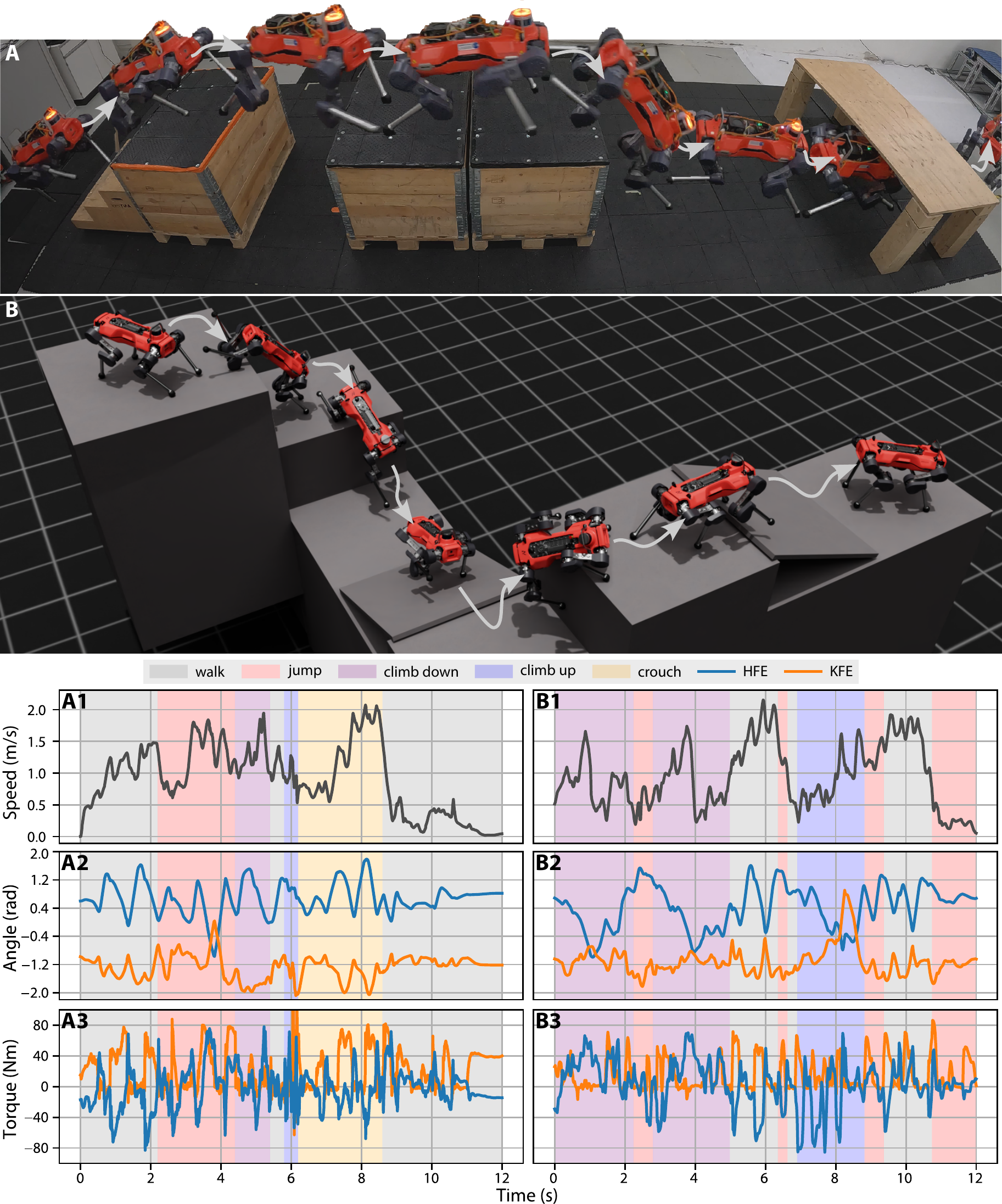}
\caption[Trajectories of the robot in simulation.]{Deployment of the pipeline on the robot ANYmal D. (\textbf{A}) Trajectory on the real robot. (\textbf{B}) Trajectory in simulation. (\textbf{A1})-(\textbf{A3}) and (\textbf{B1})-(\textbf{B3}) depict the profiles of the robot's speed, the selected skills, and two joint angles and torques corresponding to (\textbf{A}) and (\textbf{B}), respectively. The system leverages the motor's full torque capabilities and uses large deflections of the joints to reach high speeds and overcome challenging obstacles.}
\label{fig:line_vel_combo}
\end{figure*}

The three learning-based modules operate together without expert demonstration, offline computation, or a priori knowledge of the environment and enable the robot to reliably reach a target across different arrangements of randomized obstacles. Fig.~\ref{fig:line_vel_combo} shows two trajectories and the corresponding profiles of the robot's speed, the selected skills, and the joint positions and torques for one of the leg's hip flexion-extension (HFE) and knee flexion-extension (KFE) motors.
The robot crosses the terrain swiftly and chooses suitable skills at every time-step. It reaches speeds of up to \SI{2}{\meter/\second} and undergoes fast accelerations and decelerations (Fig.~\ref{fig:line_vel_combo} (A1) and (B1)). 
The system leverages a large portion of the motor's range and often reaches maximum torque. Along trajectory A, the HFE motor deflects by more than \SI{160}{\degree} (Fig.~\ref{fig:line_vel_combo} (A2)), which is necessary for the leg to reach the other side of the gap and catch the fall of the robot during the climb down maneuver. In trajectory B, the policy saturates the motor during the climb to propel the robot onto the \SI{0.9}{\m} high platform (Fig.~\ref{fig:line_vel_combo} (B3)).\\
The system is able to control the robot precisely despite the high speeds. In scenario A, the robot reaches the leftmost box after the stairs with a speed of \SI{1.5}{\meter/\second}. With a width of \SI{0.8}{\m}, the box is smaller than the robot's footprint in standing configuration. At this location, it has to perform precise foothold placement to pass the last step and prepare for the jump, despite the out-of-distribution scenario for the jumping skill, which has been trained with boxes double the size. This shows that the low-level skills can cope with more intricate scenes than what they have been trained on with our method.\\
In scenario B, the skill selection scheme of the navigation module is non-trivial. At several locations along the path, it chooses skills that have not been designed for the specific setting at hand. Indeed, it favors the jumping skill to quickly turn the robot on the spot in the narrow passages after the first step down or before the climb. This can be explained by the jumping skill's training set-up, where it has to jump from one box to another, and the initial and target headings are randomized. The skill learns to turn on the spot in tight spaces and is more capable in such scenarios compared to other skills. The navigation module is able to discover such strengths during its training process and exploits them on deployment.
It is worth mentioning that the switches of the low-level skills are smooth and unnoticeable on the real robot.

The pipeline is also able to recover from disturbances or crashes. We show that the robot stands up and completes the course after falling down from a box, and that it can pass a table after heavily slipping due to low ground friction. Moreover, the system is able to quickly readapt its trajectory when obstacles are pulled away from the robot during execution, despite the fact that all the components are trained with static environments only. This is due to the fast reaction times of each component, and the ability of the perception module to quickly correct its output when there is a mismatch between its belief state and the current measurements.

In the following analysis, we delve into each component of our proposed approach, revealing how such behaviors can be effectively achieved.

\subsection{Locomotion Module}
\begin{figure*}[t!]
\centering
\includegraphics[width=\textwidth]{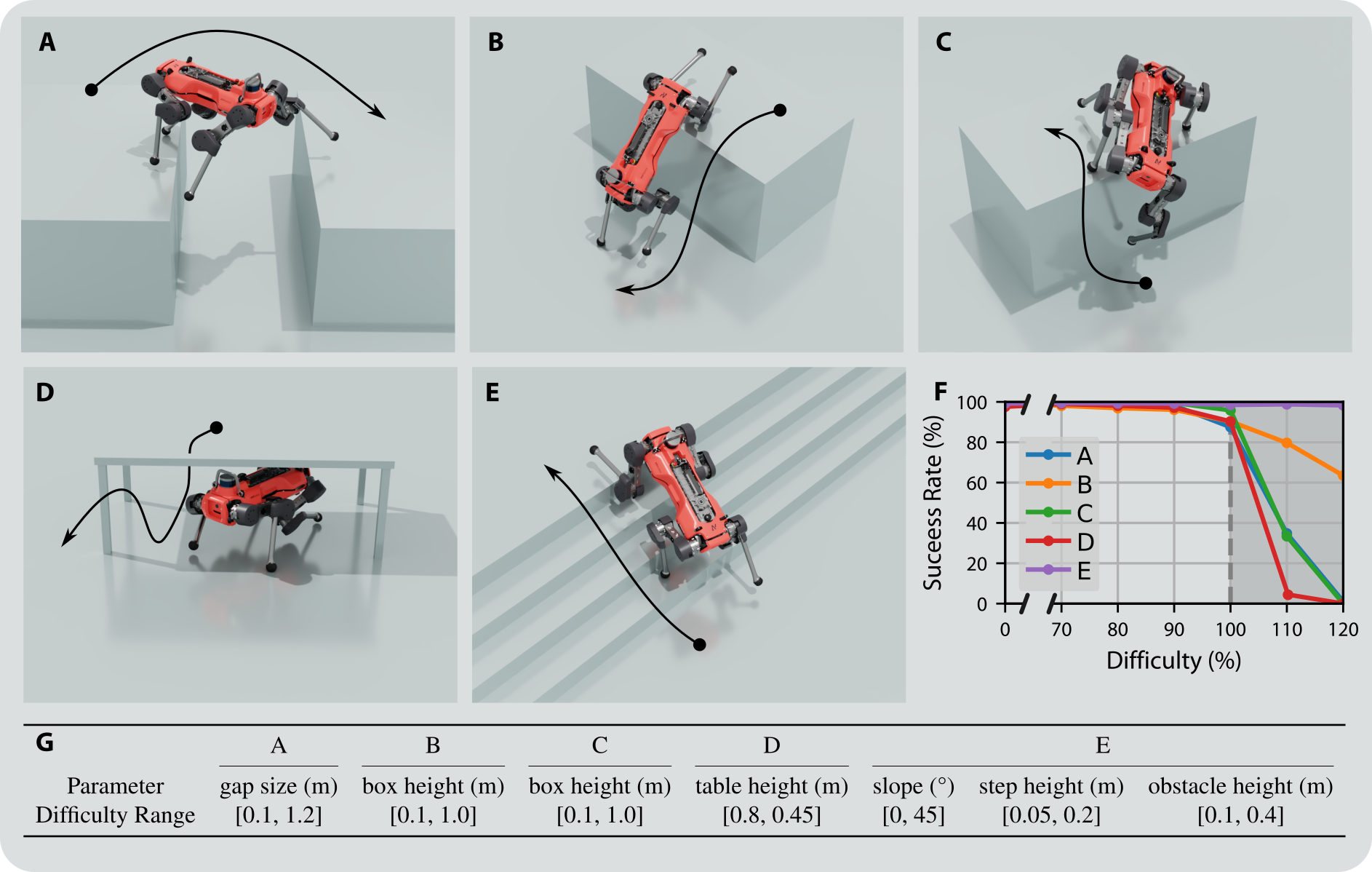}
\caption[Locomotion skills.]{Training scenarios of the locomotion skills with the resulting behaviors. 
(\textbf{A}) Jumping. (\textbf{B}) Climbing down. (\textbf{C}) Climbing up. (\textbf{D}) Crouching. (\textbf{E}) Walking. (\textbf{F}) Success rate of each skill for obstacles of varying difficulty. (\textbf{G}) Ranges of parameters used during training (0\% to 100\% in \textbf{F}).}
\label{fig:locomotion_skills}
\end{figure*}

First, we analyze the performance and emerging behavior of each locomotion policy separately. In Fig.~\ref{fig:locomotion_skills}, we show the training setup with the corresponding learned behavior of each skill and evaluate the performance of the policies across obstacles of increasing difficulty.

\paragraph{Jumping}
The robot starts on a box and must jump to a neighboring box separated by a gap of up to \SI{1}{m}. 
In order to perform a successful jump, the robot approaches the gap sideways and carefully places its feet as close as possible to the edge before using the full actuation power to leap to the other side. It uses three legs to propel itself, while the fourth is extended to land on the other side. The robot then transfers two diagonal legs before bringing the last leg across the gap. Due to randomization, the policy keeps the feet at a safe distance from the edge and can recover from missteps and slippage by transferring the robot's weight between the non-leaping legs.

\paragraph{Climbing Down}
The robot starts on a box with a height of up to \SI{1}{m} and must climb down to reach a target on the ground. 
Since we penalize high impacts on its feet to prevent motor damage, the robot first goes on its knees on the edge and brings its center of gravity as low as possible.
It then jumps down to land on its front legs holding its weight with the back knees on top of the box. It then takes a few steps forward on the front legs to re-position itself and allow the back legs to come down gently. 
The policy learns to be robust to small shifts in the perceived terrain by slowly pushing its feet over the edge until it makes contact with its knees. It then uses the conveniently L-shaped shank and knees of the robot as a hook on the edge of the box.  

\paragraph{Climbing Up}
The robot starts on the ground and must climb on top of a box with a height of up to \SI{1}{m}. 
To climb to the top, the robot puts one of its front feet on the top surface and uses it to lift itself to an upright configuration. It then re-positions itself before jumping to land with a hind leg on the top while balancing by pushing against the vertical surface with the fourth leg. Finally, it propels the whole body up and brings the fourth leg on top. 
While the robot only uses its feet and shanks when possible, it also learns to use its knees when needed. For example, if the third leg slips or misses the edge, the robot can use its knee to recover without falling back to the ground.

\paragraph{Crouching}
The robot must reach a target located on the other side of a narrow passage with a minimum height of \SI{0.4}{m}. 
When it crosses a table, the robot adopts the expected behavior of lowering its base while walking in the desired direction. With a low base height, it must adapt its gait and use both hip motors to lift its feet off the ground.

\paragraph{Walking}
The robot must traverse various irregular terrains consisting of stairs, slopes, and randomly placed small obstacles. These diverse terrains are traversable with a common walking policy and are similar to the terrains used in previous perceptive legged locomotion works \cite{miki2021locomotion, rudin2022minutes}. 
The policy is capable of scaling and descending short slopes of \SI{40}{\degree}, climbing steps of \SI{0.25}{m} step height, and running on flat ground at \SI{2}{\meter/\second}.
Due to the diversity of training scenarios, this policy generalizes well to unseen terrains such as narrow stairs, slopes, or combinations of different obstacles. 

Fig. \ref{fig:locomotion_skills} (F) shows the success rate of each skill across a range of corresponding obstacles with increasing difficulty. The displayed range covers 0\% to 120\% of the maximum obstacle difficulty during training. All skills perform well up to 90\% of their respective difficulty. After that, the crouching skill's performance drops the quickest when the passage becomes narrower than the height of the robot. The performances of jumping, climbing, and climbing down skills also drop sharply due to the physical limits of the robot. Finally, the walking skill extrapolates well beyond the training range of difficulties since the corresponding terrains are less challenging.
  
\subsection{Navigation Module}

We examine the emerging behaviors of the navigation module and show that it exhibits the following desired characteristics:

\begin{enumerate}
\item Terrain-adapted path selection: The planner is able to select sub-goals based on its instantaneous measurement of the terrain by extracting 3D information from the latent space of the perception module. For similar environment configurations, it adapts the path depending on the obstacles' dimensions.
\item Low-level policy switching and control: The high-level module selects the most appropriate policy based on the terrain and can send the right commands to control the robot's trajectory. It takes into account the capabilities of each skill.
\end{enumerate}

\begin{figure*}[pt!]
\centering
\includegraphics[width=\textwidth]{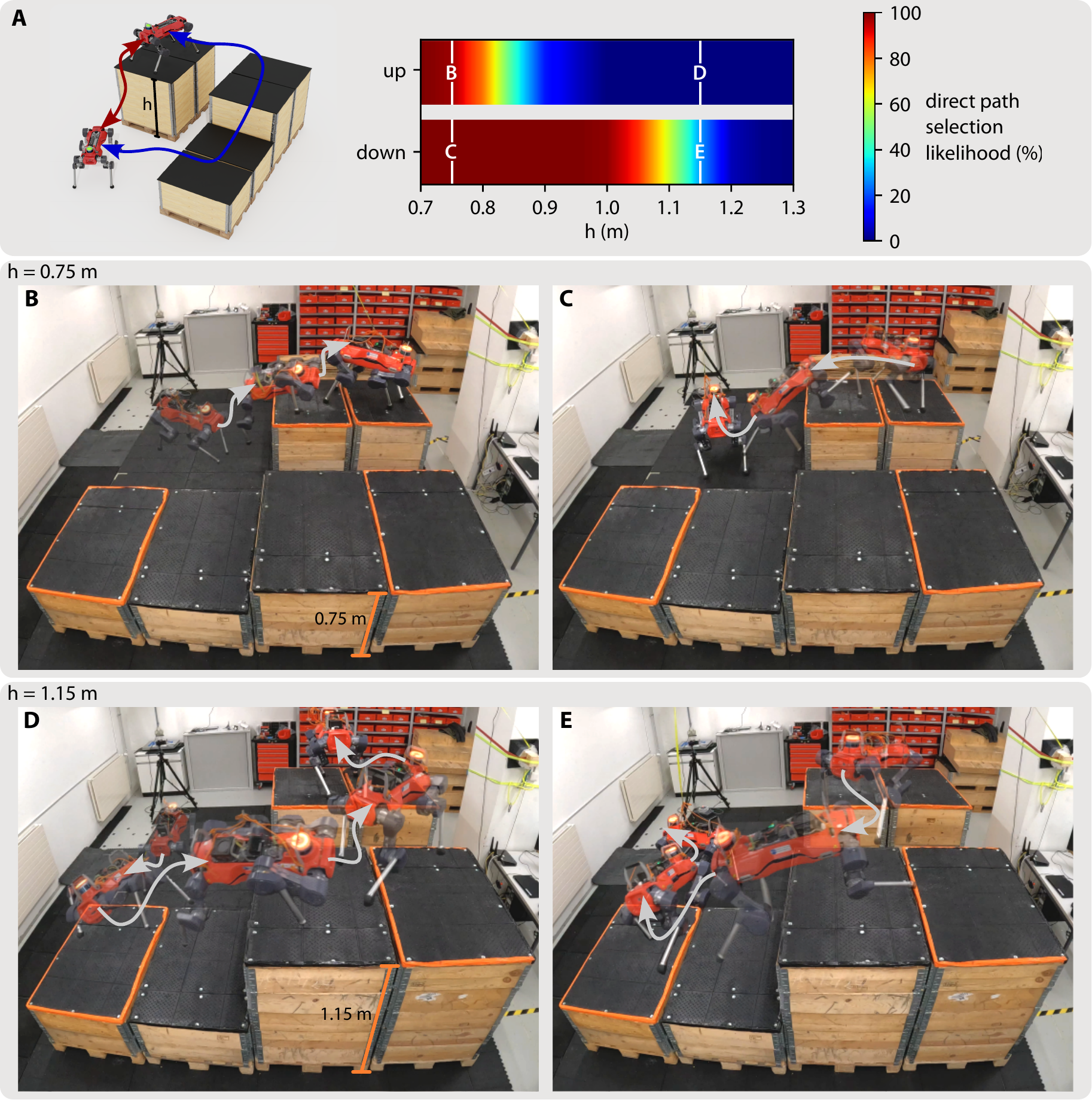}
\caption[Adaptive path selection]{Adaptive path selection. The robot starts on the ground and is given a target on top of the box in the back, and then commanded back to the initial position. (\textbf{A}) Likelihood of going up and down along the direct path (red line) as a function of the height of the box. (\textbf{B}) and (\textbf{C}) Deployment on the robot for h = \SI{0.75}{\meter}. (\textbf{D}) and (\textbf{E}) Deployment on the robot for \mbox{h = \SI{1.15}{\meter}}. For the same targets and box placement, the navigation policy chooses a different path depending on the height of the boxes to reach the goal.
}
\label{fig:segments}
\end{figure*}

Upon convergence, the navigation policy can fully control the five locomotion skills across the course to solve the problem (Fig.~\ref{fig:line_vel_combo} and Fig.~\ref{fig:segments}). This task is not trivial due to the position-based formulation these policies are trained with. Indeed, each low-level policy can modulate the robot's movement freely within the allocated time and must only comply with the position and heading commands when the time is over. For example, it could track the orientation command at any time along the trajectory. Therefore, the navigation policy has to learn how to properly combine the position, heading, and timing commands for each skill to achieve the desired motion of the robot. This is particularly important when the robot arrives at high speeds on a narrow obstacle. It often has to quickly decelerate the robot and then turn on the spot to get to the next obstacle. 

The navigation module is aware of the capabilities and limitations of each skill and uses this knowledge to adapt the trajectory. This is primarily visible with the climb up, climb down, and crouch skills, where depending on the configuration of the obstacle, it will modify its output.
When a box is too high, the policy does not go up or down directly since it would result in failure. For tables that are too low, it will climb over them rather than crouch underneath.
Such adaptation is depicted in Fig.~\ref{fig:segments}, where the robot starts on the ground and we command the policy to reach the target box in the back (up) and then command it back to the starting position (down). In (A), we show the likelihood that the robot takes the direct path as a function of the height of the box. It can be seen that when the height of the box increases, the policy is more likely to choose a longer but safer path. Indeed, the policy sends the robot down directly until a height of \SI{1}{\meter}, after which it increasingly prefers to take the longer route. On the other hand, it switches much more quickly to the longest route when going up. This difference can be explained by the different disturbances we add during high-level training, which have a stronger impact on the climb up skill.
(B) and (C) show the resulting trajectories on the real robot for \hbox{h = \SI{0.75}{\meter}}, and (D) and (E) for h = \SI{1.15}{\meter}.

Another example where the robot has to distance itself from the target to reach distant goals is described in supplementary section~S4.



We compare the performance of our method against a manually computed trajectory  (Table~\ref{tab:comparison}) for the different terrains depicted in Fig.~\ref{fig:worlds}. For the manual trajectory, we hard-code the commands and skills along the course based on the sequence of obstacles, which amounts to human expert demonstrations. 
The study is performed on three randomly selected scenarios with 1000 roll-outs each, where the obstacles' difficulty is close to the maximum defined during low-level training (100\% in Fig.~\ref{fig:locomotion_skills}).
The table shows that manually placing targets performs well in certain scenarios, but fails in other cases where the locomotion policies require finer-grained control.
Moreover, our high-level policy learns to dynamically adjust the targets by placing them further away to increase the speed of the robot. Manual demonstrations with targets at key locations (i.e. in the middle of obstacles) lead to lower speeds thus requiring a longer time to reach the target.
Finally, human demonstrations do not scale well when randomizing the terrain, since it requires hand labeling each new case.

We provide an ablation study of the policy's action space in supplementary section~S5.

\begin{table}
    \centering
    \caption{Comparison of the navigation policy's performance against a manually hard-coded trajectory.}
    \begin{tabular}{c c c}
        \toprule
        { } & Ours & \makecell{Manual} \\
        Terrain: Fig. \ref{fig:worlds} - A & 98.2\% & 95.3\% \\
        Terrain: Fig. \ref{fig:worlds} - B & 96.3\% & 60.9\% \\
        Terrain: Fig. \ref{fig:worlds} - C & 97.6\% & 75.3\% \\
        \bottomrule
    \end{tabular}
    \label{tab:comparison}
\end{table}
 
\subsection{Perception module}

\begin{figure*}[pt!]
\centering
\includegraphics[width=0.86\textwidth]{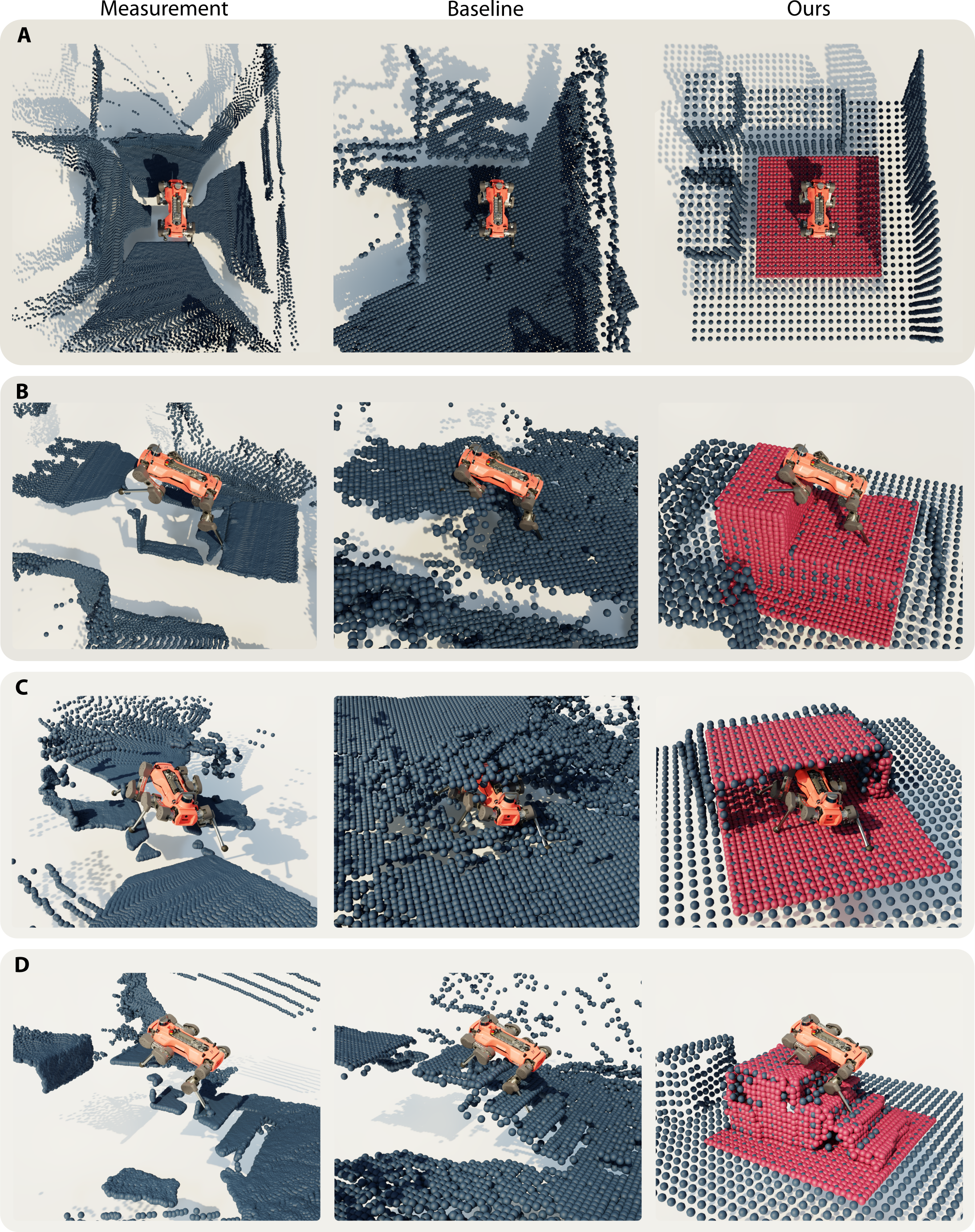}
\caption[Terrain reconstructions for different scenarios]{Terrain reconstructions for different scenarios (real-world data). The first column shows the point cloud measurements, the second the baseline elevation map~\cite{miki2022EM} viewed as a point cloud, and the last corresponds to the reconstruction with our method. For our method, we show the coarse-resolution output in blue and the high-resolution output (refinement process) in red.}
\label{fig:reconstruction}
\end{figure*}

The perception module can process the noisy and occluded point cloud measurements to produce a meaningful latent for the navigation module and a clean reconstruction for the locomotion module. As mentioned earlier, the module operates asynchronously with the rest of the system on deployment. This differs from our training set-up, where we assume that the perceptive information and the resulting reconstruction and latent are available at the exact time of inference. The performance does not seem to be affected by such delays in the real system. 

We analyze the reconstructions and compare them against an elevation mapping baseline \cite{miki2022EM} that runs alongside our network. This method provides several improvements to the commonly used framework described in \cite{Fankhauser2018ProbabilisticTerrainMapping}, making it a stronger contender for the parkour task. It can detect drift in z-direction to realign the map to the correct height and performs additional visibility checks to remove outliers. While we mainly qualitatively evaluate the reconstruction performance, we refer the reader to~\cite{hoeller2021DeepNav} for a quantitative analysis for this type of approach.

Several outputs of the network for scenarios Fig.~\ref{fig:segments} (D) and Fig.~\ref{fig:line_vel_combo} (A) are presented in Fig.~\ref{fig:reconstruction} (real-world data). The first column corresponds to the measurements, the second to the baseline map visualized as a point cloud, and the last to our reconstruction. Since the baseline is an elevation map, the corresponding point cloud does not contain vertical surfaces. Our approach produces a multi-resolution output and we color the high-resolution output (refinement process \SI{2}{\meter} around the robot) in red, and the coarse-resolution output in blue for better distinction. Note that the coarse-resolution output (blue points) within the red regions is only shown for comprehension and is not used by the rest of the pipeline.

From the various outputs, it can be seen that the network is able to cope with sparse measurements and correctly estimate the layout of the scene. In (A), the points falling on the edge of the boxes are used as evidence to reconstruct the upper parts at the right height. The surface on the right of the robot is correctly identified as a wall and reconstructed accordingly. On the other hand, the baseline does not consider the regions on top of the higher boxes since no measurements have yet reached these locations.

The coarse network produces less precise reconstructions further away from the robot due to the lower resolution of the voxels and noisy measurements along some of the obstacles' edges. In (A), for example, while the estimated height of the box to the left of the robot is correct, the width is approximately 8 cm too large. However, nearby the robot, the refiner can deal with such inaccuracies and further enhances the reconstruction. This can be seen in (D), where the refiner produces cleaner stairs than the coarse map.

The importance of the auto-regressive feedback can be witnessed when the robot crouches under the table in (C). Despite the sparsity of the measurements on the top surface, the network remembers this region since it could be seen during the approach in previous time steps. Of course, the baseline method is not designed to handle such scenarios with overhangs. It produces a mix containing the top surface at some locations and the ground at others, resulting in an erroneous map.

The robustness to state estimation drift can be seen in (B) and (D) by comparing with the baseline. In (B), the robot's position estimate suddenly jumped to the left. Our network detects such situations and immediately corrects the map. The elevation map, on the other hand, cannot cope with the drift and the knees of the robot and the hind leg are inside the map. The same happens in (D), where the hind leg is inside the elevation map.

\section{Discussion}
This work aims to extend the capabilities of legged robots on highly challenging terrains. We have presented a complete pipeline for robotic parkour, including specially developed low-level locomotion skills, a high-level navigation module, and a perception module. The proposed approach allows the robot to move with unprecedented agility. It can now evolve in complex scenes where it must climb and jump on large obstacles while selecting a non-trivial path toward its target location. The dynamic nature of the task poses multiple challenges that render existing approaches unsuitable. It requires non-standard locomotion skills at the actuation limit, a planner with an intrinsic understanding of the locomotion capabilities with respect to the surrounding obstacles, and a perception module capable of inferring the three-dimensional topology of the terrain based on the partial observations provided by the sensors. 

We propose a fully learned approach where each module employs one or multiple neural networks. The networks are trained in simulation and transferred to the real world. We demonstrate that our task can be solved without pre-mapping or offline planning, and all required computations can happen onboard the robot in real-time.
Using learning-based modules is advantageous for real-world deployment. The complexity of solving the task is shifted to the learning stage. Once the relatively small networks are trained, they display complex behaviors at almost no cost compared to optimization or sampling-based methods, without resorting to limiting assumptions or simplifications.

\subsection{Current Limitations}
The pipeline has some limitations that remain to be tackled for deployment in realistic and unstructured scenarios. First, the scalability of the method to more diverse scenarios remains to be tested. We showcase the system's capabilities in a limited range of scenarios, utilizing a handful of distinct modules within the environment. In order to scale to complex environments such as a collapsed building or even a real parkour course would require the robot to perceive, navigate, and cross a wider variety of obstacles. While we can always train more low-level skills, provide more data to the perception module and train the navigation module in more diverse scenarios, it remains to be seen how well these different modules can generalize to completely new scenarios.
 
Furthermore, training the whole pipeline can be time-consuming since it uses a total of eight neural networks, each requiring separate tuning. Some of them are interdependent, meaning that modifying one requires retraining the others. For instance, the navigation module can only receive the latent tensor of the specific perception module it was trained on and has to use the same locomotion policies. In turn, the perception module needs to be re-trained if a skill adopts a different motion or if a new obstacle is introduced. Simultaneous training of the different components might be necessary in the future.

Finally, since the navigation module must make a series of correct decisions to reach the goal with many possibilities leading to failure, the algorithm requires many iterations to converge. We develop a specific curriculum to overcome this limitation. Without this step, the robot struggles to discover the correct behaviors and gets stuck in front of larger obstacles. A possible solution would be to pre-train the navigation module using expert demonstrations, for example by finding candidate solutions with brute-force search.

\section{Materials and Methods}

\subsection{Overview}

\begin{figure*}[t!]
\centering
\includegraphics[width=0.9\textwidth]{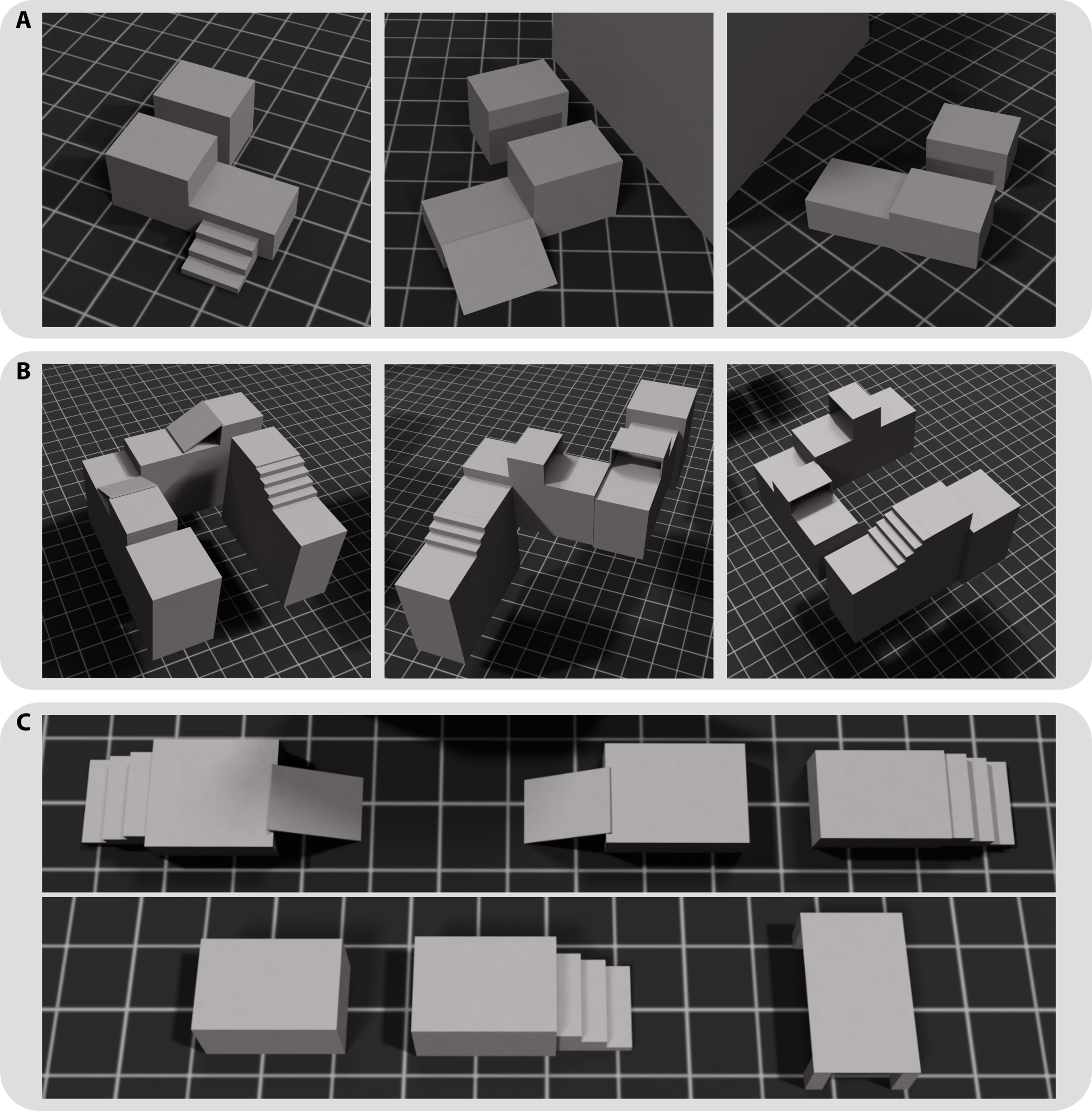}
\caption[Types of environments used for training.]{Types of environments used for training. The dimensions of the individual obstacles and the arrangements are randomized.}
\label{fig:worlds}
\end{figure*}

The goal of the agent is to navigate and locomote in an environment to reach a specific target location within a short amount of time. 
We constrain the task to different configurations of pallet-sized boxes, allowing us to keep the main challenges of agile navigation while having a feasible, structured, and repeatable scenario.

We create three different terrain types presented in Fig.~\ref{fig:worlds}:
\begin{enumerate}
  \item (A) Different arrangements of boxes, where the robot might have to climb and jump over a gap to reach the target. The dimensions of these elements and the target's and robot's initial position are randomized. The robot must reach any of the boxes starting from the ground or reach a target on the ground starting from one of the boxes. Depending on the setting, the robot can either reach the target directly, or it might need to go around the environment to find a lower box to climb on first. Walls and distracting objects are added to increase the generalization of the perception and navigation modules to realistic scenarios.

\item (B) A parkour line consisting of a long winding platform with multiple obstacles on the way. The robot must traverse the obstacles without falling off the platform. The shape of the platform, the sequence of obstacles, and their parameters are randomized.
\item (C) A simplified version of the parkour line for real-world deployment. Instead of a winding platform, the obstacles are arranged in a straight line on the ground, and the robot is not allowed to walk around them. Again, the sequence and parameters of obstacles are randomized. We also add walls and distracting objects next to the line.
\end{enumerate}

Each terrain displays different capabilities of the pipeline. Scenario A demonstrates the general applicability to realistic but relatively constrained scenarios. The navigation module has to understand the capabilities of the locomotion skills and choose the path accordingly. Even though the obstacles arrangements are fairly constrained, the robot can start anywhere on the terrain and must choose different paths depending on the target location and obstacle parameters.
On the other hand, scenario B shows generalization to more randomized scenarios with different platform shapes and obstacle arrangements. While there is only one possible path, the sequence of obstacles leads to various cases that the navigation and perception modules must learn to handle correctly. Finally, scenario C allows us to force the robot to climb on the obstacles without having to recreate a high winding platform with gaps on either side for real-world deployment. Due to the different formulation, we use a separate navigation policy for that scenario.

\subsection{Pipeline}

The pipeline consists of three learning-based modules, which are described in the following subsections. Supplementary sections~S1 and S2 define the observations, actions, and rewards of the locomotion and navigation policies and provide further implementation details.

\subsubsection{Perception Module}

The perception module plays a crucial role for the downstream pipeline and endows the robot with scene understanding. The navigation and locomotion modules both use its output to make path planning, policy selection, foothold placement, and contact decisions.
It ingests point clouds of the scene coming from depth cameras and LiDAR to produce an estimate of the terrain around the robot. The measurements from these sensors are noisy and heavily occluded by obstacles in the environment or the robot itself.

To overcome these challenges, we opt for a data-driven method with an encoder-decoder architecture inspired by~\cite{hoeller2022TREP}. However, here, we develop a multi-resolution scheme that consists of two networks operating at different scales, see Fig.~\ref{fig:architecture}. It allows us to balance the trade-off between reconstruction accuracy and map size. Indeed, close to the robot, the map is smaller and has a higher resolution since this region is essential for locomotion. Further away, the resolution is lower, allowing for a broader view of the scene. The navigation module in these areas only needs the approximate configuration of the scene for path planning and policy selection, making a lower resolution sufficient.

The encoder takes in the point cloud and compresses it into a compact representation. The decoder uses this representation and generates an output that completes the missing information and filters out the noise.
Additionally, the coarse-resolution network benefits from an auto-regressive feedback, where the previous output is transformed into the current frame and concatenated with the measurement. This allows the module to accumulate evidence over time and reconstruct the scene's elements that are no longer visible. For example, when the robot passes below a table, the module can use the aggregated information from previous frames to estimate the layout of the table and reconstruct the top surface, even if it is currently not visible to the sensors. This is also necessary with certain maneuvers, such as climbing, where the robot's limbs often block a large portion of the left and right cameras, see supplementary section~S6.

The measurements are first converted to a voxel grid around the robot. In each occupied voxel, a feature describes the position of the centroid of the points that fall within that voxel. The features of unoccupied voxels are set to 0. Dense 3D convolutions are performed over the dense voxel grid. While the authors in \cite{hoeller2022TREP} use a sparse implementation, it does not scale well with the reinforcement learning set-up with 4000 robots. Surprisingly, the dense formulation can handle such a large batch size with sufficient speeds but this comes at the cost of high memory requirements (approximately 45 GB GPU memory).

The decoder outputs the voxel occupancy probability as well as the position of the centroid for each cell. The reconstructed point cloud can then be recovered by pruning the cells whose occupancy probability is below a user-defined threshold.
Contrary to \cite{hoeller2022TREP}, we do not use skip connections to produce a more informative latent that the navigation module can directly use. While this might limit the generalization performance, we found that it works well for our task with randomized parkour worlds.

The high-resolution network uses the features of the coarse-resolution network's last layer as input, along with the point cloud measurements. Note that it does not use an auto-regressive feedback, since temporal information is already contained in its input.

As mentioned earlier, the goal of the coarse-resolution network is to provide a broad view of the scene. Therefore, we use a voxel size of \SI{12.5}{\cm}, resulting in a map of \SI{4}{\m} along each axis of the robot. The high-resolution voxels have a size of \SI{6.25}{\cm} resulting in a map size of \SI{2}{\m}.

We train these networks in an unsupervised fashion from simulated data on a total of 2000 trajectories with 100 time-steps each. We equally split the data set across the different parkour scenarios.
The occupancy output is trained using a binary cross-entropy loss, while the centroids are trained using the Euclidean distance to the ground truth. We follow the same data augmentation procedure described in \cite{hoeller2022TREP}. It consists of perturbing the position of the points, adding random blobs, removing patches of points, and noisifying the robot's position. As we show in the results, this is key to make the pipeline robust to noise and drift.

\subsubsection{Locomotion Module}
The locomotion module is an interface that exposes the low-level skills to the rest of the pipeline and operates at \SI{50}{Hz}.
It contains a catalog of policies, each trained for a specific locomotion skill: walking, climbing up, climbing down, crouching, and jumping.
These skills are trained using reinforcement learning and output joint position commands for the motors.
The module receives a signal indicating which skill to activate. 

As input, the policies receive the current proprioceptive state, a local map of the surrounding terrain, an intermediate command, and output position commands to the motors.
The skills are trained separately and share the observation and action spaces but require different flavors of rewards and termination conditions in order to be trained efficiently. The training set-up closely resembles~\cite{rudinAdvancedSkills} and uses position-based commands. Instead of tracking velocity commands, the robot must reach a target position within a given time. In addition to the position and time commands, we add a heading target, specifying the yaw orientation the robot must adopt by the end of the trajectory. Furthermore, we implement symmetry augmentations and find that they solve the asymmetry issues reported in \cite{rudinAdvancedSkills} and lead to more robust policies. We describe this procedure in the supplementary section~S3.
While the navigation module receives a full 3D representation of the map, it is impractical for the locomotion policies due to their high update rate and the corresponding computational cost during training. We resort to using a 2.5D elevation map around the robot, which can directly be computed from the point cloud output of the perception module. To bridge the reality gap, we perturb the elevation map during training by adding noise to individual points and shifting the map up to \SI{7.5}{\cm} in all directions. This forces the policies to adopt a safer behavior and encourages robustness to slight imperfections in the map reconstructions.

Below, we delineate the various skills and, if applicable, the modifications made to the training configuration:

\paragraph{Walking} The robot must traverse various irregular terrains consisting of stairs, slopes, and randomly placed small obstacles, similar to the ones commonly used in previous legged locomotion works \cite{miki2021locomotion, rudin2022minutes}. 

\paragraph{Jumping} The robot starts on a box and must jump to a neighboring box separated by a gap of up to \SI{1}{\m}. We use a curriculum on the size of the gap.

\paragraph{Climbing down} The robot starts on a box with a height of up to \SI{1}{\m} and must climb down to reach a target on the ground. We use a curriculum on the height of the box. We add a termination condition on high impact forces on the feet. This termination is essential to get a transferable motion. Without it, the robot learns to jump down from the top, which is possible in simulation but leads to potential damage on the real robot. 

\paragraph{Climbing up} The robot starts on the ground and must climb on top of a box with a height of up to \SI{1}{\m}. We use a curriculum on the height of the box. We allow the robot to make contact with the base and knees by reducing the weights of the corresponding penalties. This leads to the natural progression where the policy first learns to climb using its knees and then starts using its feet instead when possible.

\paragraph{Crouching} The crouching policy has the specificity of dealing with overhanging obstacles. The robot must reach a target located on the other side of a narrow horizontal passage with a minimum height of \SI{0.4}{\m}. We use a curriculum on the height of the passage. We provide the same 2.5D map as the other policies. As such, it sees the obstacle from the top and cannot differentiate a table from a box. This does not pose a problem since it is only trained in such scenarios and will always try to go under the obstacle.

While the walking policy is trained on a mix of terrains (60\% stairs, 20\% slopes, and 20\% randomized obstacles), the other specialized skills are all trained with 80\% of their corresponding obstacle and 20\% of random rough terrain. This leads to more natural gaits and better performance upon deployment.

\subsubsection{Navigation Module}
The navigation module guides the robot around the terrain to reach the target within the allocated time. 

The network is trained in a hierarchical set-up using reinforcement learning. It consists of an outer loop running the navigation policy at \SI{5}{Hz} and an inner loop running the locomotion module at \SI{50}{Hz}. The locomotion policies of the inner loop are frozen throughout training. At every high-level time step, the navigation policy receives the relative position of the final goal, the remaining time to accomplish the task, the robot's base velocity, orientation, and the latent tensor of the perception module. It then selects a locomotion skill and guides the latter with a local position, heading, and time command.
Similar to the training of locomotion policies, we employ the time-dependent command formulation described in~\cite{rudinAdvancedSkills}. The agent is given a fixed time to reach the goal, and the distance-to-goal penalty is only activated on the last time-step of the episode. This sparse formulation allows the policy to explore the terrain to find safer paths and take its time where needed. The episode is also terminated if the robot falls or the contact forces are too high. 
To speed up convergence, we employ a curriculum where we first place the global targets close to the robots' starting positions and then move them further away on the terrain as the reward increases.

To accommodate for the formulation, we modify the PPO algorithm and augment the actor's multilayer perceptron with a hybrid output. The last layer's features are split to form a Gaussian distribution for the commands and a categorical distribution for skill activation. The categorical distribution assigns a selection probability for each of the low-level skills.
During training, the actions are sampled from the respective distributions to enable exploration. On deployment, we use the mean of the Gaussian and select the policy with the highest assigned probability.

Compared to other approaches such as \cite{hoeller2021DeepNav}, which deploy simplified kinematic models in the inner loop, rolling out the actual low-level policies during training is necessary to perform agile navigation. Indeed, the agent can make informed decisions taking into account 
the mode of operation, the capabilities, and the limitations of each low-level controller.
It can infer when a box is too high to climb on and first move towards a lower one. It carefully places the target on narrow passages to enable fine-grained foot placement. It favors the climb-down policy on lower boxes, to step down to avoid high contact forces. \\
Since the low-level skills are trained with the position-based formulation, the navigation policy must carefully combine and adjust the time, position, and heading commands to achieve the desired motion. 

\section{Acknowledgments}
\textbf{Funding}
The project was funded by NVIDIA, the Swiss National Science Foundation (SNF) through the National Centre of Competence in Research Robotics, the European Research Council (ERC) under the European Union’s Horizon 2020 research and innovation program grant agreement No 852044 and No 780883. The work has been conducted as part of ANYmal Research, a community to advance legged robotics.


\bibliographystyle{IEEEtran}

\IEEEtriggeratref{35} 
\bibliography{references}

\clearpage
\setcounter{table}{0}
\makeatletter 
\renewcommand{\thetable}{S\@arabic\c@table}
\makeatother

\setcounter{figure}{0}
\makeatletter 
\renewcommand{\thefigure}{S\@arabic\c@figure}
\makeatother

\section*{Supplementary materials}
\subsection*{S1. Observations, actions, and rewards definitions}\label{app:obs_and_rews}
\begin{table}[H]
\centering
\caption{Symbols.}
\label{tab:symbols}
\begin{tabular}{rl}
\hline
\textbf{Symbol}      & \textbf{Description}       \\ \hline
$\mathbf{r}$, $\mathbf{r}^*$ & Current and local target base positions \\
$\psi$, $\psi^*$ & Current and local target base headings \\
$t^*$  & Remaining time to reach the local target \\
$\mathbf{r}_G^*$ & Global target position \\
$t_G^*$  & Remaining time to reach the global target \\
$\alpha$ & Angle between base z-axis and gravity \\
$\mathbf{v}_b$, $\boldsymbol{\omega}_b$ & Base linear and angular velocities in base frame \\
$\mathbf{g}_b$ & Gravity vector in base frame \\
$\mathbf{q}, \dot{\mathbf{q}}$, $\dot{\mathbf{q}}_{\lim}$ & Joint positions, velocities, and velocity limits \\
$\mathbf{q}^*, \mathbf{q}_d$ & Desired and default joint positions\\
$\boldsymbol{\tau}$, $\boldsymbol{\tau}_{\lim}$ & Joint Torques and torque limits \\
$\mathbf{v}_f$, $\mathbf{F}_f$ & Feet linear velocity and contact force \\
$\mathbf{h}$ & $\SI{2}{\meter}\times\SI{1}{\meter}$ grid of height measurements around the robot \\
$\mathbf{l}$ & Scene belief state (perception module latent tensor) \\
$s$ & Index of the selected locomotion skill \\
$\mathds{S}_L$ & Target reached (locomotion)\\
{} & $ \mathds{S}_L=\mathds{1}_{\lVert \mathbf{r}_{xy} - \mathbf{r}^*_{xy}\rVert < 0.25} \mathds{1}_{\lVert \psi - \psi^* \rVert < 0.5}$\\
$\mathds{S}_N$ & Target reached (navigation)\\ 
{} & $\mathds{S}_N=\mathds{1}_{\lVert \mathbf{r} - \mathbf{r}_G^*\rVert < 0.4}$\\
\hline\\
\end{tabular}
\end{table}

\begin{table}[H]
\centering
\caption{Locomotion Rewards.}
\label{tab:reward}
\begin{tabular}{rcl}
\hline
\textbf{Reward Term}      & \textbf{Expression}  & \textbf{Weight}     \\ \hline
Position tracking &$ \mathds{1}_{t^*<1}(1 - 0.5\lVert\mathbf{r}_{xy}-\mathbf{r}_{xy}^*\rVert)$  & 10\\ 
Heading tracking & $ \mathds{1}_{t^*<1}(1 - 0.5\lVert\psi-\psi^*\rVert)$  & 5\\ 
Joint velocity & $\lVert \dot{\mathbf{q}} \rVert ^2$ & -0.001 \\ 
Torque & $\lVert \boldsymbol{\tau}\rVert^2$ & -0.00001\\ 
Joint velocity limit & $\sum_{i=1}^{12} \max (\lvert\dot{\mathbf{q}}_{i}\rvert-\dot{q}_{\lim},0)$ & -1  \\ 
Torque limit & $ \sum_{i=1}^{12} \max (\lvert\boldsymbol{\tau}_{i}\rvert-\tau_{\lim},0)$ & -0.2 \\ 
Base acc. &$\lVert \dot{\mathbf{v}} \rVert ^2 + 0.02\lVert \dot{\boldsymbol{\omega}} \rVert ^2$ & -0.001 \\ 
Feet acc.   &$ \sum_{f=1}^4 \lVert \dot{\mathbf{v}}_{f}\rVert$ & -0.002 \\ 
Action rate   &$\lVert \mathbf{q}^*_t-\mathbf{q}^*_{t-1} \Vert ^2$ & -0.01 \\ 
Feet contact force  & $\sum_{f=1}^4 \max(\lVert F_{f} \rVert - 700, 0)^2$ & -0.00001 \\ 
Don't wait & $ \mathds{1}(\lVert \mathbf{v}_b \rVert < 0.2)$ & -1 \\ 
Move in direction & $ \cos \langle \mathbf{v}_b, \mathbf{r}^*-\mathbf{r} \rangle$  & 1\\ 
Stand at target & $\mathds{S}_L \lVert \mathbf{q} - \mathbf{q}_d\rVert$ & -0.5\\ 
Collision & $\mathds{1}_{\textit{knee/shank collision}}$ & -1 \\ 
Stumble & $\mathds{1}_{\lVert F_{f,xy}\rVert > 2\lVert F_{f,z}\rVert} $ & -1 \\
Termination & $\mathds{1}_{\textit{base collision}} + \mathds{1}_{F_f>1500}$ & -200\\
\hline
\end{tabular}
\end{table}

\begin{table}[H]
\centering
\caption{Navigation Rewards.}
\label{tab:reward_hl}
\begin{tabular}{rcl}
\hline
\textbf{Reward Term}      & \textbf{Expression}  & \textbf{Weight}     \\ \hline
Position tracking &$ \mathds{1}_{t^*=0}(40\mathds{S}_N - \lVert \mathbf{r} - \mathbf{r}_G^*\rVert) $  & 0.15\\ 
Termination & $\mathds{1}_{\alpha<\pi/2} + \mathds{1}_{F_f>2500}$ & -0.5\\ 
\hline
\end{tabular}
\end{table}

\begin{table}[H]
\centering
\caption{Locomotion and Navigation Observations.}
\label{tab:observations}
\begin{tabular}{rcc}
\hline
 \textbf{Observation} & \textbf{Locomotion}  & \textbf{Navigation}     \\ \hline
$\mathbf{v}_b$ & $\times$ & $\times$ \\
$\boldsymbol{\omega}_b$ & $\times$ & {} \\
$\mathbf{g}_b$ & $\times$ & $\times$ \\
$\mathbf{q}, \dot{\mathbf{q}}$ & $\times$ & {} \\
$\mathbf{r}^*, t^*, \psi^*$ & $\times$ & {}\\
$\mathbf{r}_G^*, t_G^*$ & {} & $\times$\\
$\mathbf{h}$ & $\times$ & {}\\
$\mathbf{l}$ & {} & $\times$ \\
\hline
\end{tabular}
\end{table}

\begin{table}[H]
\centering
\caption{Navigation and Locomotion Actions}
\label{tab:actions}
\begin{tabular}{rc}
\hline
 \textbf{Module} & \textbf{Action}     \\ \hline
Locomotion &$\mathbf{q}^*$\\
Navigation & $s$, $\mathbf{r}^*, t^*, \psi^*$\\
\hline
\end{tabular}
\end{table}


\subsection*{S2. Implementation details}\label{app:impl_details}
We train all the policies and collect the data for the perception module using the Isaac Gym simulator~\cite{makoviychuk2021isaac}, where we deploy 4096 agents in parallel. To generate and prepare the perceptive data for the perception network, we develop custom CUDA kernels using Warp~\cite{macklin2022warp}. At every time-step, these kernels perform raycasting for the six depth cameras and the LiDAR for each robot ($\approx$ 140 million rays total) and directly convert them to the voxel grid inputs without copying memory. 
It is worth mentioning that on the real robot, the point cloud messages of the six Realsense cameras reach the perception node with a delay of up to \SI{250}{\milli\second}, which is prohibitively long for fast maneuvers such as climbing. Therefore, we disabled point cloud publishing for these cameras and directly subscribe to the depth image messages instead, reducing the delay to \SI{25}{\milli\second}. We project the images to point clouds and merge them with the LiDAR measurements within the node. 
In the following, we provide the rewards for the policies.

\subsection*{S3. Symmetric data augmentation for locomotion training}\label{app:symmetry}
\begin{figure}[ht]
\centering
\includegraphics[width=0.8\linewidth]{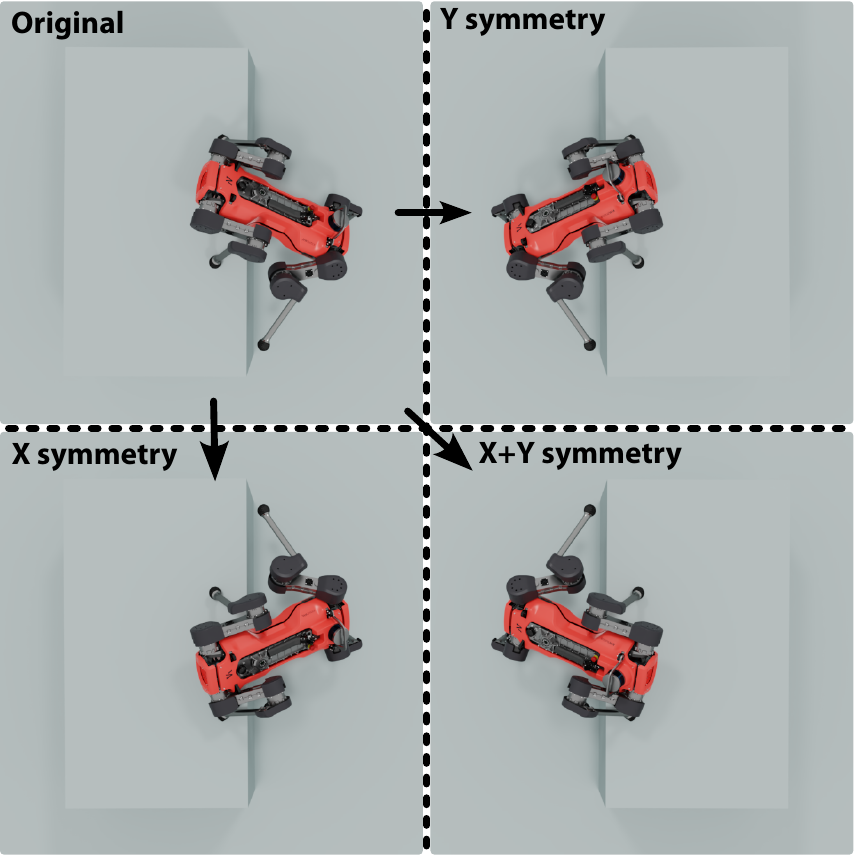}
\caption[Representation of Symmetry augmentation]{Representation of the symmetric state augmentation. The original state (top-left) is augmented into four symmetric states using the X and Y symmetries of the robot.}
\label{fig:symmetry}
\end{figure}

The position-tracking formulation of the locomotion training does not constrain the robot's trajectory between the initial and target positions. This allows the robot to learn complex behaviors but also leads to asymmetric motions. For example, the climbing policy only learns to climb forwards and prefers to turn the robot around if facing an obstacle backward, leading to unfavorable situations when the robot must cross multiple obstacles with different skills. We solve these issues by exploiting the symmetric nature of the robot. Based on the duplication method of \cite{abdolhosseini2019learning}, we augment each environment transition with all symmetric variants by transforming the observations and actions accordingly. Specifically, we use front-back and left-right symmetries of the ANYmal D robot.

The authors in \cite{abdolhosseini2019learning}, however, mention that their duplication method suffers from poor convergence due to the off-policy nature of the mirrored states.
Indeed, these augmentations result in low probabilities for the transformed actions for not fully trained policies. We resolve this issue by setting the probability of the original actions to all symmetric variants. Intuitively, we bootstrap the learning process of a randomly initialized policy since we know that at convergence symmetric states will lead to symmetric actions with equal probability.

\subsection*{S4. Navigation across long ranges}\label{app:navigation_ushape}
\begin{figure}[t]
\centering
\includegraphics[width=\linewidth]{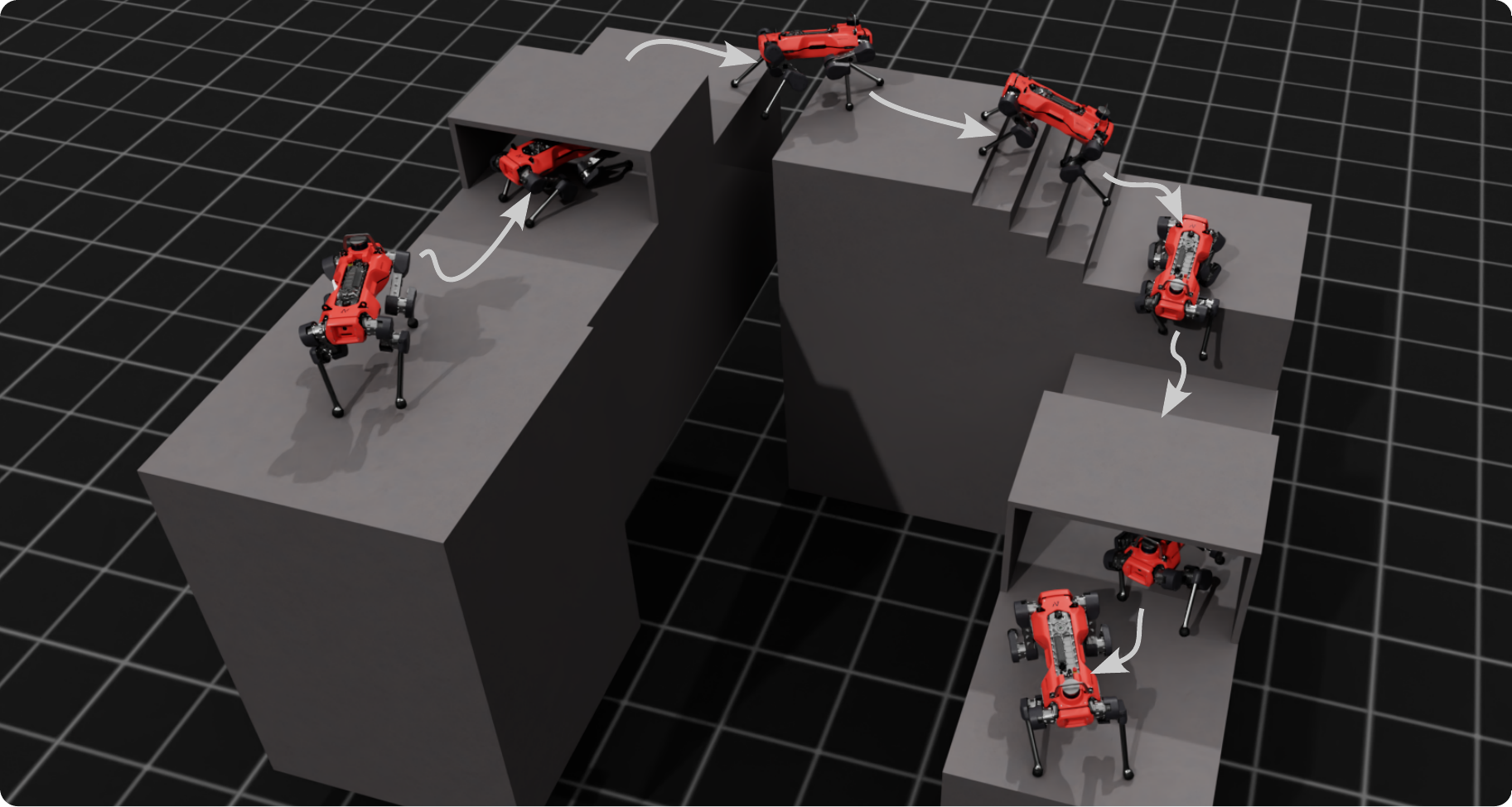}
\caption[Emergent high-level behaviors]{The planner can reach remote targets, even if it has to distance itself from the goal first.}
\label{fig:ushape}
\end{figure}

Fig.~\ref{fig:ushape} depicts a scenario with a distant goal in a U-shaped terrain. The planner understands that it cannot cross the wide gap by jumping, and it must first distance itself from the target to solve the task.

\subsection*{S5. Ablation study of the navigation module's output}\label{app:navigation_ablation}

In Table~\ref{tab:ablation}, we analyze the importance of the different components of the navigation policy's action space. We remove the timer output and set a fixed time for the low-level policies (No T); we remove the heading output and set it to be in the direction of the next target (No H); we remove both (No H, No T).
The study is performed under the same conditions as the comparison with the manually coded trajectory (Table~\ref{tab:comparison}).
The results show that using the heading and time commands for the low-level skills increases the performance of the system. Specifically, we can see that adding the heading command increases the success rate on terrains where the robot must quickly turn multiple times on the spot (Terrain Fig.~\ref{fig:worlds} (A)), while the time command leads to better performance in longer terrains (Terrain Fig.~\ref{fig:worlds} (B) \& (C)), where the high-level policy must use fast motions on simple obstacles, but slow down in risky parts.

\begin{table}[h!]
    \centering
    \caption{Comparison of the navigation policy's performance against different formulations. (\textbf{No T}): no time output. (\textbf{No H}): no heading output. (\textbf{No H, No T}): no heading or time output.}
    \begin{tabular}{c c c c c}
        \toprule
        { } & Ours & \makecell{No T} & \makecell{No H} & \makecell{No H, No T} \\
        Terrain: Fig.~\ref{fig:worlds} - A & 98.2\% & 94.7\% & 89.5\% & 81.1\% \\
        Terrain: Fig.~\ref{fig:worlds} - B & 96.3\% & 89.4\% & 88.1\% & 71.9\% \\
        Terrain: Fig.~\ref{fig:worlds} - C & 97.6\% & 91.6\% & 94.6\% & 94.0\% \\
        \bottomrule
    \end{tabular}
    \label{tab:ablation}
\end{table}

\subsection*{S6. Description of the measurement blind spots}\label{app:blind_spots}
\begin{figure}[ht]
\centering
\begin{subfigure}{.25\textwidth}
  \centering
  \includegraphics[width=\linewidth]{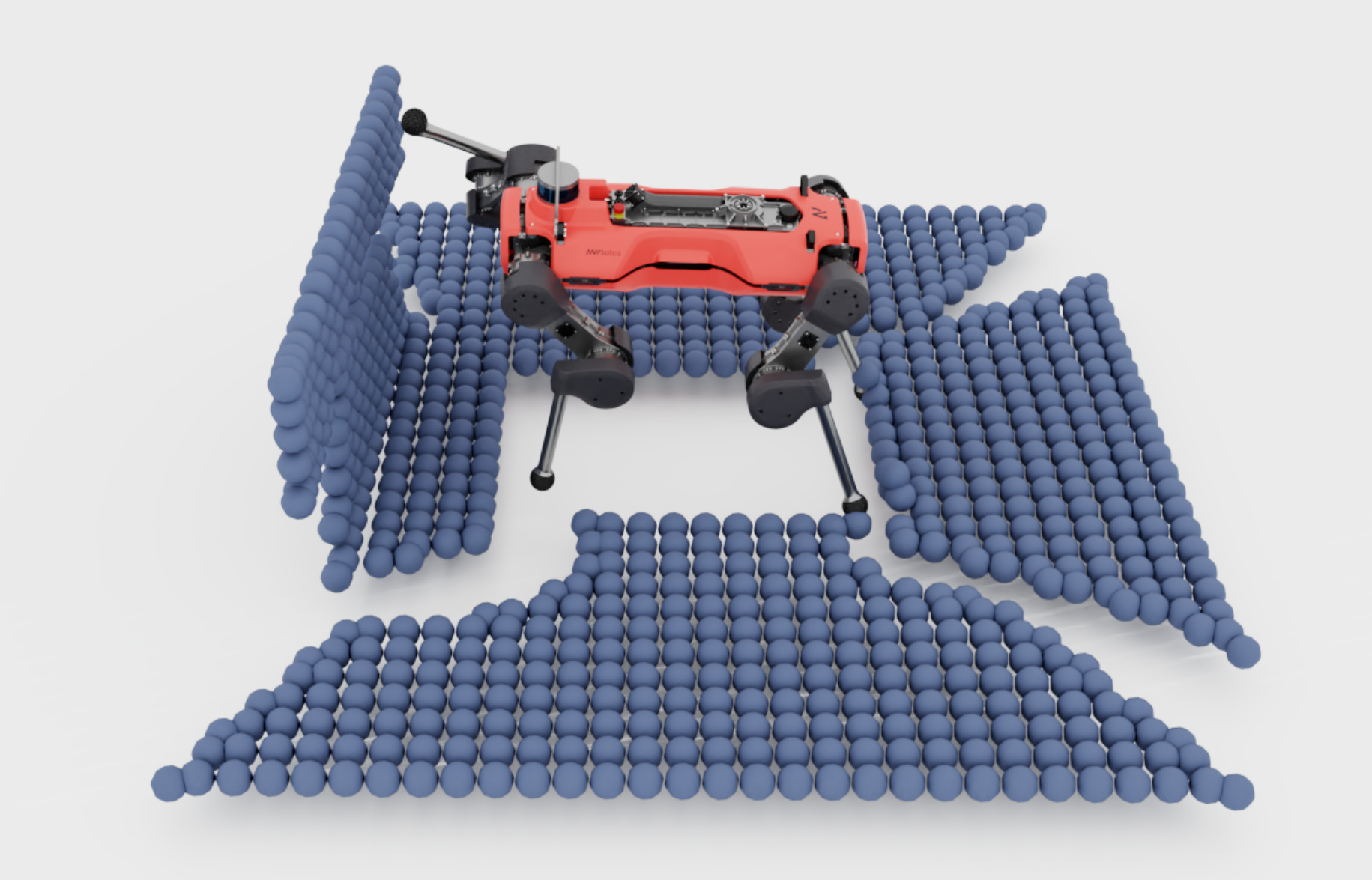}
\end{subfigure}%
\begin{subfigure}{.25\textwidth}
  \centering
  \includegraphics[width=\linewidth]{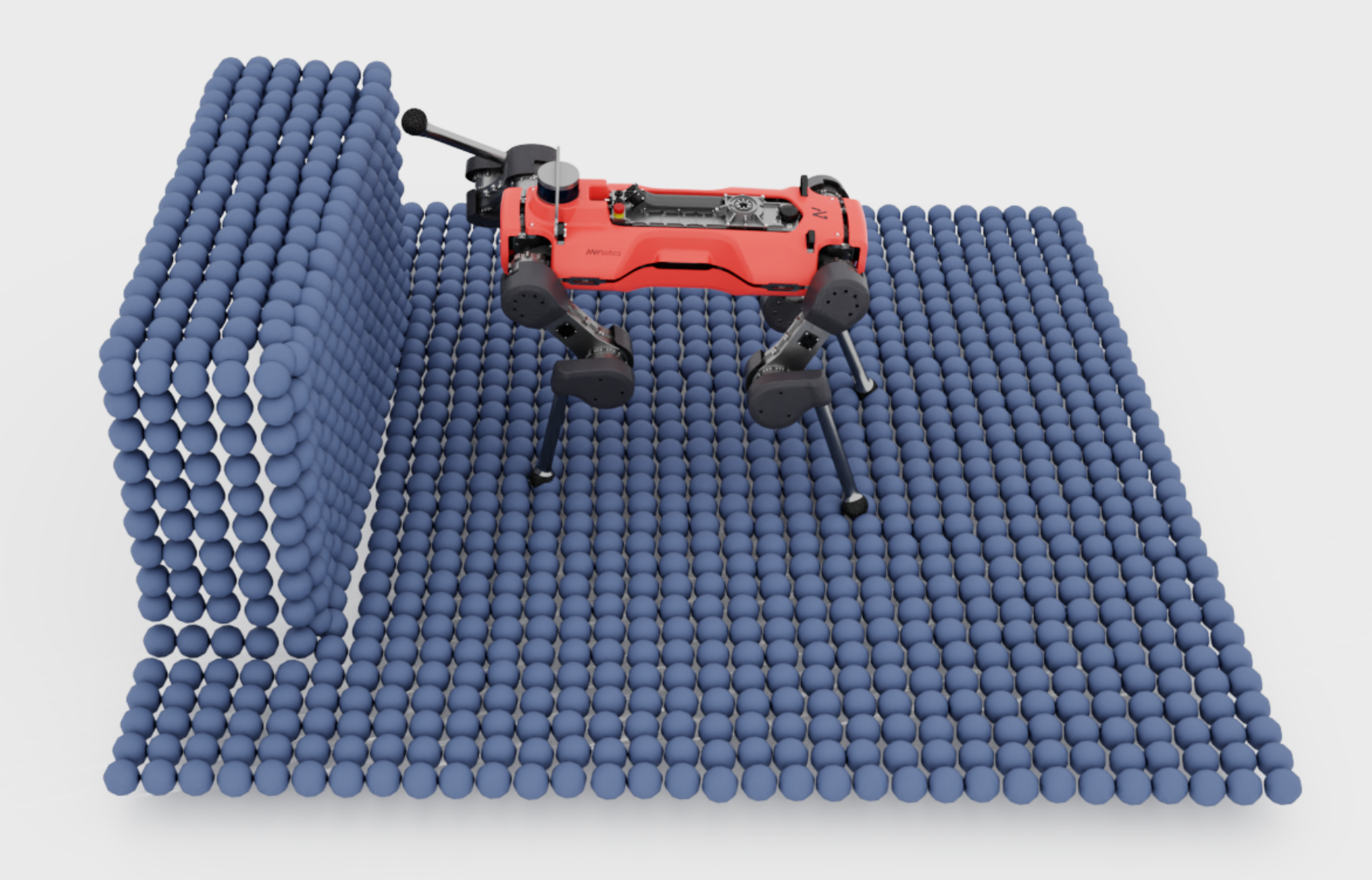}
\end{subfigure}
\par \vspace{1mm}
\begin{subfigure}{.25\textwidth}
  \centering
  \includegraphics[width=\linewidth]{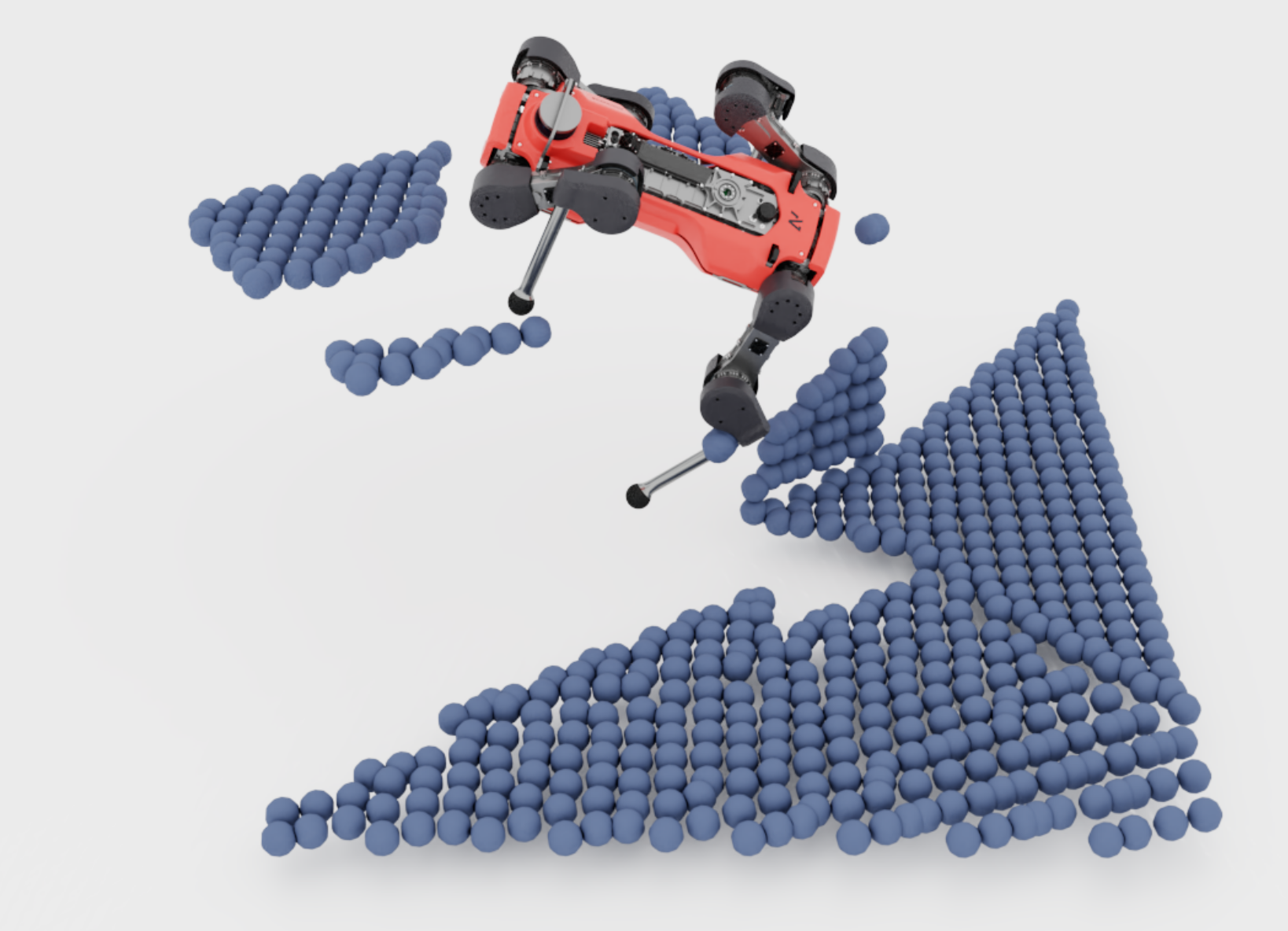}
  \caption{Measurement}
\end{subfigure}%
\begin{subfigure}{.25\textwidth}
  \centering
  \includegraphics[width=\linewidth]{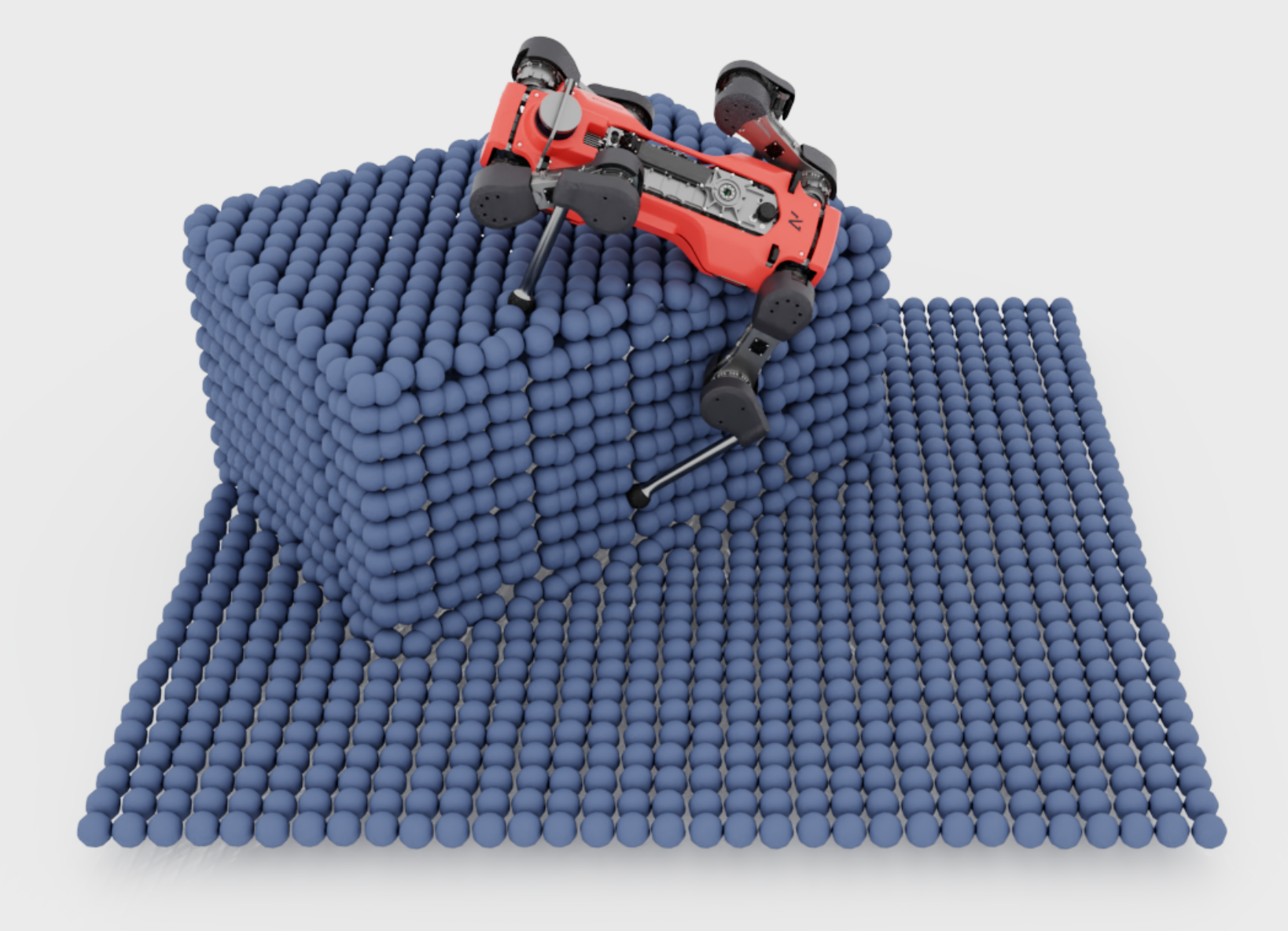}
  \caption{Ground truth}
\end{subfigure}
\caption[Measurement blind spots occurring during a box climbing maneuver]{Measurement blind spots occurring during a box climbing maneuver. The sensors do not perceive the top surface at the beginning (top row). The perception module has to use the points on the edge of the front surface to estimate the height of the box and correctly reconstruct the top. During the climb (bottom row), the limbs obstruct the cameras resulting in large occluded areas.}
\label{fig:climb_pointclouds}
\end{figure}

Measurement blind spots during a box climbing maneuver can be seen in Fig.~\ref{fig:climb_pointclouds}. Due to the height of the box, the robot cannot perceive the top surface at the beginning. During the climb, large occluded regions occur because of limb obstructions. It can also be seen that the camera arrangement is not particularly favorable for locomotion since there are blind spots immediately below the robot.

\subsection*{S7. Incorrect terrain reconstructions}\label{app:incorrect_reconstructions}

\begin{figure}[ht]
\centering
\includegraphics[width=\linewidth]{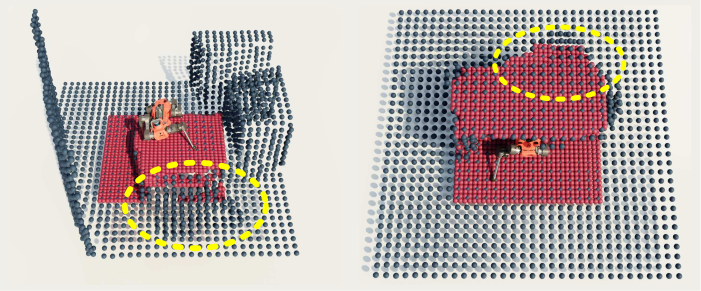}
\caption[Incorrect reconstructions with our method.]{Incorrect reconstructions with our method, highlighted in yellow. On the left, the network hallucinates a stair. On the right, it inflates the shape of the table.}
\label{fig:wrong_reconstruction}
\end{figure}

There are situations where the network produces wrong outputs (Fig.~\ref{fig:wrong_reconstruction}). When climbing on the first box (left), the network tends to hallucinate a stair behind it in the measurement blind spot, probably due to a data-set imbalance. This does not impede the performance of the navigation and locomotion modules since the network quickly corrects this erroneous output once it has a better view of the situation. Also, the table is sometimes inflated when the robot crawls underneath (right). This comes from a combination of measurement sparsity on the top surface and state estimation drift, which is more pronounced for crouching maneuvers. Again, this does not pose a problem to complete the task, since the robot would stay crouched for longer in the worst case. 

\end{document}